\documentclass{article}

\usepackage[table]{xcolor}

\usepackage[final]{corl_2026}

\usepackage{graphicx}
\usepackage{booktabs}
\usepackage{multirow}
\usepackage[utf8]{inputenc}
\usepackage{newunicodechar}
\newunicodechar{，}{,}

\usepackage{amsfonts}
\usepackage{nicefrac}
\usepackage{microtype}
\usepackage{bbding}
\usepackage{amsmath}
\usepackage{amssymb}
\usepackage{mathtools}
\usepackage{bm}
\usepackage{tcolorbox}
\usepackage{soul}
\usepackage{subcaption}
\usepackage{wrapfig}
\usepackage{float}

\usepackage[accsupp]{axessibility}

\definecolor{szu_color}{HTML}{84193E}
\definecolor{cvprblue}{rgb}{0.21,0.49,0.74}
\definecolor{hku_color}{HTML}{13A983}
\definecolor{color1}{RGB}{255,250,205}
\definecolor{color2}{RGB}{255,228,225}
\definecolor{color3}{RGB}{34,139,34}

\def\logo{\makebox[22pt][l]{\raisebox{-0.9ex}{\includegraphics[height=20pt]{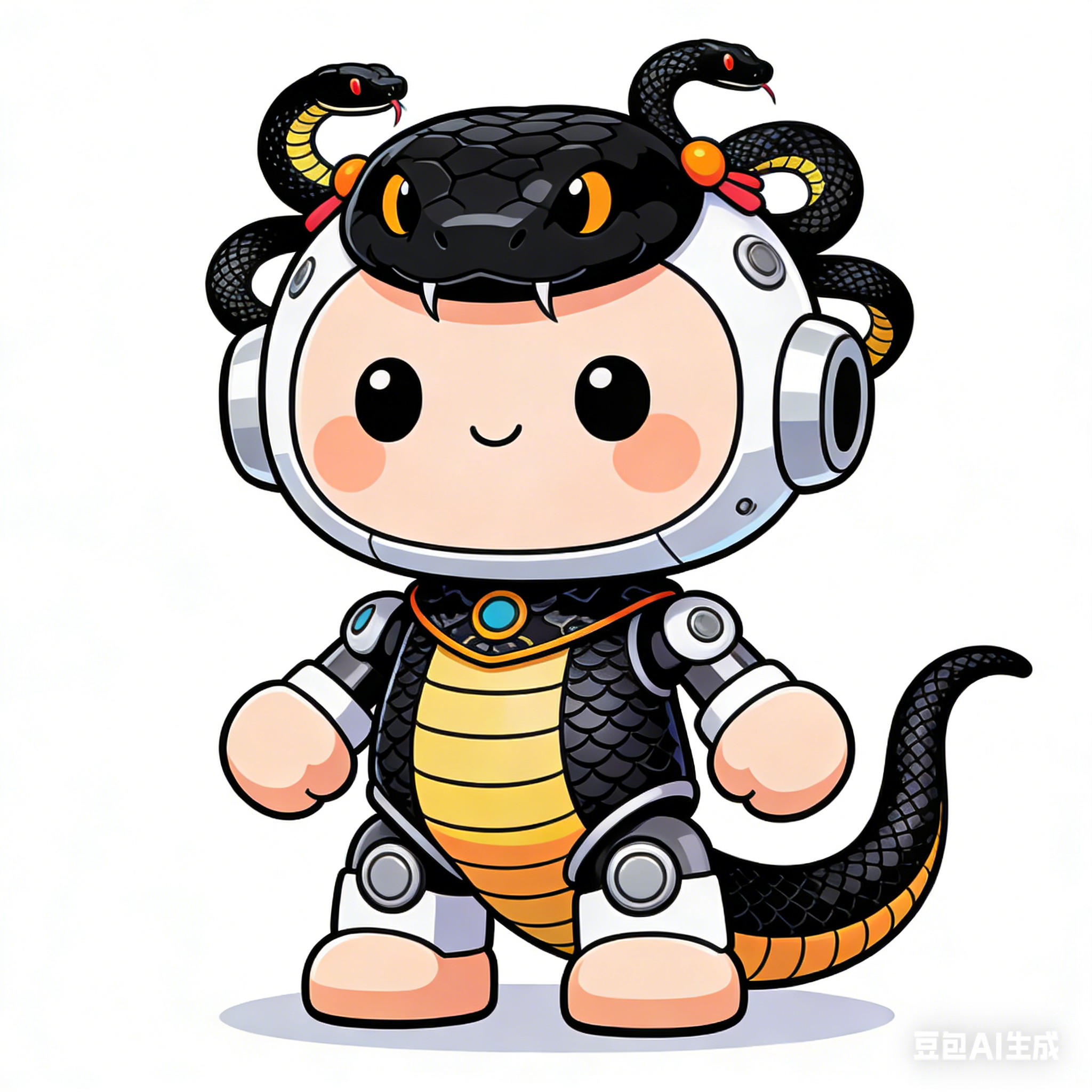}}\hspace{15pt}}}

\title{RoboDreamer: Anticipatory Humanoid Locomotion with Predictive State-Space Models}

\author{
  Zhe Li$^{*,\dagger}$\\
  Mars Lab, NTU\\
  \And
  Yangyang Wei$^{*}$\\
  BAAI\\
  \And
  Xichen Yuan\\
  Mars Lab, NTU\\
  \And
  Zhenzhe Zhang\\
  PKU\\
  \And
  Weihao Yuan\\
  NJU\\
  \And
  Shanghang Zhang\\
  PKU\\
  \And
  Jianfei Yang$^{\ddagger}$\\
  Mars Lab, NTU\\
}
\begin{document}
\maketitle
\renewcommand{\thefootnote}{\fnsymbol{footnote}}

\footnotetext[1]{Equal Contribution.}
\footnotetext[2]{Project Lead.}
\footnotetext[3]{Corresponding Author.}

\begin{figure}[h]
    \centering
    \includegraphics[width=0.8\linewidth]{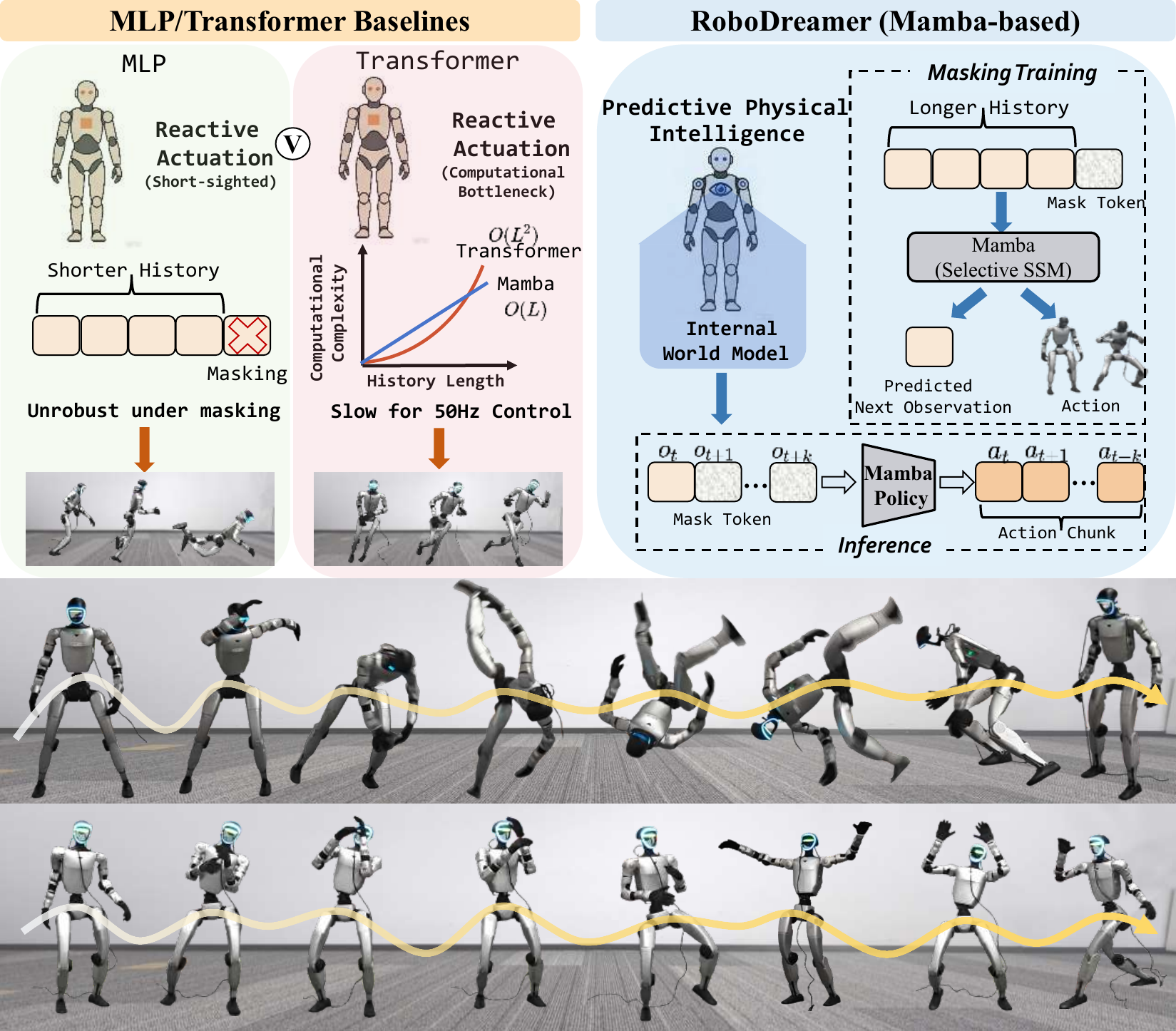}
    \caption{\logo \textbf{RoboDreamer} studies anticipatory humanoid tracking under imperfect sensing. Randomized temporal masking trains the policy to infer missing recent observations from history; the same masking interface is reused at inference for closed-loop refinement and optional action chunking. Mamba serves as the temporal backbone rather than the primary methodological contribution.}
    \label{fig:motivation}
\end{figure}
\begin{abstract}
Humanoid locomotion requires control policies that remain stable under imperfect sensing while exploiting temporal context for consistent motion. We present RoboDreamer, a two-stage teacher--student framework that combines next-observation consistency with randomized continuous temporal masking. A teacher is first trained on clean observations, and a student is then distilled under masked recent observations, encouraging the policy to infer missing current information from history. At inference, the same masking interface is reused for implicit closed-loop action refinement and optional multi-step action chunking. Mamba is used as the temporal backbone, while matched ablations show that masking/distillation provides a substantial part of the gain and Mamba contributes additional tracking improvements with real-time latency. Experiments in IsaacLab, MuJoCo, and on a Unitree G1 demonstrate robust motion tracking under observation masking and successful real-world deployment.
\end{abstract}


\section{Introduction}
\label{sec:intro}

Human locomotion is not merely a sequence of reactive motor responses, but a continuous process that exploits temporal context to maintain a coherent estimate of the underlying physical state. When sensory information is temporarily missing or unreliable, humans do not immediately lose balance; instead, recent motion history and physical continuity provide sufficient cues to infer the current state and anticipate its near-future evolution. Humanoid control, however, is still commonly formulated as a largely reactive mapping from recent observations to motor commands. Such policies can perform well under clean sensing, yet become brittle once the observation stream is interrupted by sensor dropouts, communication latency, packet loss, or transient corruption. More fundamentally, a controller that only reacts to the currently available observation has limited ability to explicitly reason about what information is missing and how the physical system is likely to evolve through that missing interval.

This limitation is particularly evident in prevalent MLP-based humanoid tracking policies~\citep{he2025asap,xie2025kungfubot,li2025language,liao2025beyondmimic,li2025you,li2025robomirror,li2025bfm}. MLPs are computationally efficient and therefore attractive for high-frequency control, but their temporal reasoning ability is limited when recent observations are partially unavailable. A missing or delayed observation can directly perturb the action prediction because the policy has no explicit mechanism for reconstructing the absent state from temporal context. Transformer-based policies~\citep{vaswani2017attention,ma2026robust} provide stronger sequence modeling through self-attention, but the quadratic cost with respect to sequence length makes long temporal contexts increasingly expensive for high-frequency humanoid control. More recently, Mamba-based controllers such as HuMam, LocoMamba, and HumanoidMamba have demonstrated that selective state-space models can serve as efficient temporal backbones for robotic control~\citep{wang2026humam,wang2026locomamba,zhou2026humanoidmamba}. These developments establish Mamba as a promising architecture for temporal control, but simply replacing an MLP or Transformer with Mamba does not by itself teach a policy how to reason under missing observations.

In this work, we instead focus on \emph{how temporal information should be modeled and supervised for robust humanoid control}. Our central hypothesis is that a controller should not only consume historical observations, but should be explicitly trained to infer missing physical states from that history and to predict how the system will evolve locally in time. Based on this idea, we propose RoboDreamer, a two-stage teacher--student framework built around masked predictive distillation. We first train a teacher policy using clean observations. A student policy is then trained from scratch through a DAgger-style distillation procedure~\citep{ross2011dagger}, while recent observations are randomly and continuously masked. Instead of directly exposing the RL optimization to severe observation corruption, the clean teacher provides stable action supervision, while the student is forced to recover the information necessary for control from its temporal context. In this way, temporal masking becomes a structured learning objective rather than merely a robustness augmentation.

To further regularize the temporal representation, we introduce a next-observation consistency objective that encourages the student to encode locally predictive dynamics. The role of this objective is complementary to temporal masking: masking teaches the policy to infer the present from incomplete recent observations, while next-observation consistency constrains the learned latent state to remain informative about near-future physical evolution. Mamba is adopted as the temporal backbone because its recurrent state provides an efficient mechanism for maintaining and updating historical information over time. Importantly, however, the contribution of RoboDreamer lies in the masked predictive training framework; Mamba provides an efficient substrate on which this temporal inference can be realized.

The same masking semantics learned during training can also be reused at inference. When the current observation token is masked, the policy must infer the current control state from its internal temporal representation. We exploit this mechanism in two forms. First, it enables implicit closed-loop action refinement: adjacent action predictions can be fused while only the current action is executed before replanning, preserving closed-loop feedback while benefiting from temporally consistent predictions. Second, by appending additional mask tokens, the policy can optionally extrapolate multiple future steps and produce short action chunks when sensing or inference frequency is reduced. This provides a unified connection between missing-state inference and short-horizon predictive control without introducing a separate planner or trajectory-generation module.

We evaluate RoboDreamer on Unitree G1 across IsaacLab, MuJoCo, and real-robot deployment. Under observation masking and sensing uncertainty, the proposed framework achieves more robust motion tracking than reactive baselines while maintaining real-time control. Controlled comparisons across MLP, Transformer, GRU, LSTM, and Mamba backbones further show that the improvement cannot be attributed to Mamba alone: masked predictive distillation contributes a substantial portion of the robustness gain, while Mamba provides an additional tracking--latency advantage, particularly for dynamic motions. The learned temporal representation also supports stable implicit refinement and short-horizon action chunking under reduced sensing frequency.

Our contributions are summarized as follows:
\begin{itemize}
    \item We introduce a two-stage masked predictive distillation framework for humanoid motion tracking, in which a clean teacher supervises a student trained under randomized continuous temporal masking. This explicitly teaches the controller to infer missing current information from temporal context rather than merely reacting to available observations.

    \item We combine temporal masking with next-observation consistency and reuse the learned masking semantics at inference to support both implicit closed-loop action refinement and optional multi-step action chunking, connecting temporal state inference with short-horizon predictive control in a unified framework.

    \item We adopt Mamba as an efficient temporal backbone and conduct controlled comparisons with MLP, Transformer, GRU, and LSTM policies under matched history and training settings. The results show that masking drives a substantial part of the robustness improvement, while Mamba provides an additional tracking--latency advantage for real-time humanoid control.
\end{itemize}
\section{Related Work}
\label{sec:related}

\subsection{Humanoid Whole-body Control}
\label{subsec:whole_body_control}

Classical model-based frameworks for whole-body control achieve high-precision execution by leveraging analytical dynamics and rigorous contact formulations; however, they are often hindered by the immense effort required for system identification and their fragility when encountering unmodeled physical phenomena~\citep{geyer2003positive,sreenath2011compliant}. While data-driven reinforcement learning (RL) has successfully mitigated these modeling burdens, which demonstrates agility in specialized tasks such as extreme terrain navigation and dynamic recovery~\citep{wang2025beamdojo,peng2021amp,li2023robust,huang2025learning,he2025learning}, these methods typically necessitate meticulous, task-specific reward tuning and often struggle to exhibit the fluid coordination characteristic.

To manage the high dimensionality of humanoid control, some researchers have proposed bifurcating upper- and lower-body objectives into decoupled policies~\citep{zhang2025falcon,li2025hold}, though this partitioning may compromise global inter-limb synchronization. Others have explored hierarchical architectures to sequence complex maneuvers~\citep{su2025hitter}, which improves modularity but frequently introduces significant design overhead and computational bottlenecks.

In contrast, whole-body motion tracking offers a paradigm shift by adopting human demonstrations as a direct, dense supervisory signal~\citep{han2025kungfubot2}. This approach effectively bypasses the requirement for handcrafted rewards while fostering the emergence of globally synchronized and expressive behaviors across diverse repertoires. By unifying disparate motor skills under a single tracking objective, this paradigm provides a scalable and principled trajectory toward synthesizing human-like humanoid intelligence.

\subsection{Humanoid Motion Tracking}
\label{subsec:humanoid_motion_tracking}

Pursuit of lifelike maneuvers via human demonstration has transitioned from isolated clip imitation to mastery of universal skill repertoires. Early milestones like DeepMimic~\citep{peng2018deepmimic} established imitation foundations through phase-based tracking and randomized initializations. To bridge the sim-to-real gap for agile skills, ASAP~\citep{he2025asap} introduced a multi-stage delta-action framework, while specialized systems such as HuB~\citep{zhang2025hub} and KungfuBot~\citep{xie2025kungfubot} attained high-fidelity reproduction of aggressive, high-dynamic motions through meticulous motion conditioning.

The evolution toward consolidated controllers was catalyzed by OmniH2O~\citep{he2024omnih2o}, which demonstrated a universal policy capable of spanning diverse motion libraries. Subsequent advancements, including ExBody2~\citep{ji2024exbody2}, refined gait expressiveness through decomposed tracking targets, while TWIST~\citep{ze2025twist} and CLONE~\citep{li2025clone} attained precision in teleoperation settings for lower-dynamic tasks. More complex strategies have also emerged, such as BumbleBee's~\citep{wang2025experts} expert-distillation pipeline and GMT's~\citep{chen2025gmt} prioritized root-velocity tracking. BeyondMimic~\citep{liao2025beyondmimic} achieves high-fidelity single-motion tracking through crafted objectives and precise system identification, which are then distilled into a unified diffusion policy for task-specific control. KungfuBot2~\citep{han2025kungfubot2} introduces an Mixture-of-Experts framework to enable a general motion tracking policy that generalizes across a diverse set of skills. BFM-Zero~\citep{li2025bfm} proposes that off-policy unsupervised RL is a viable approach to train a behavioral foundation model for whole-body control of a humanoid robot. Sonic~\citep{luo2025sonic} achieves strong tracking performance but demands large-scale datasets and substantial computational resources. Besides, several modality-driven locomotion methods~\citep{li2025robomirror,li2026w1,xie2026textop}, including LangWBC~\citep{shao2025langwbc}, RoboGhost~\citep{li2025language}, and RoboPerform~\citep{li2025you}, rely on MLP-based tracker policies. Additionally, Sonic~\citep{luo2025sonic} achieves strong whole-body control performance by training on an extremely large-scale dataset. However, it is also based on an MLP architecture. Recently, RGMT~\citep{ma2026robust} utilizes Transformer~\citep{vaswani2017attention} as the backbone of the tracker policy, which can fuse extended temporal contexts. However, its self-attention mechanism incurs a quadratic computational cost that escalates rapidly with history length, hindering real-time deployment on high-frequency humanoid systems.

In contrast, state-space models like Mamba~\citep{gu2024mamba} offer a principled alternative, providing $O(L)$ scaling and $O(1)$ inference efficiency without sacrificing long-range temporal reasoning. Our framework, RoboDreamer, departs from fragile kinematic mimicry toward visually grounded performance control, enabling humanoids to function as semantically aligned and physically resilient imitators.
\paragraph{Temporal policy backbones and predictive control.}
Concurrent work has increasingly adopted Mamba/selective state-space models for locomotion and humanoid control. HuMam~\citep{wang2026humam} uses Mamba for state-centric humanoid RL, LocoMamba~\citep{wang2026locomamba} studies Mamba-based vision--proprioception fusion for locomotion, and HumanoidMamba~\citep{zhou2026humanoidmamba} combines Mamba with next-action prediction. OmniXtreme~\citep{wang2026omnixtreme} instead uses flow matching to scale high-dynamic multi-motion control. These works establish temporal sequence modeling as an important direction; our contribution is orthogonal, focusing on masked predictive distillation and on reusing the same masking mechanism for inference-time refinement and chunking.

\begin{figure*}[!t]
    \centering
    \includegraphics[width=\textwidth]{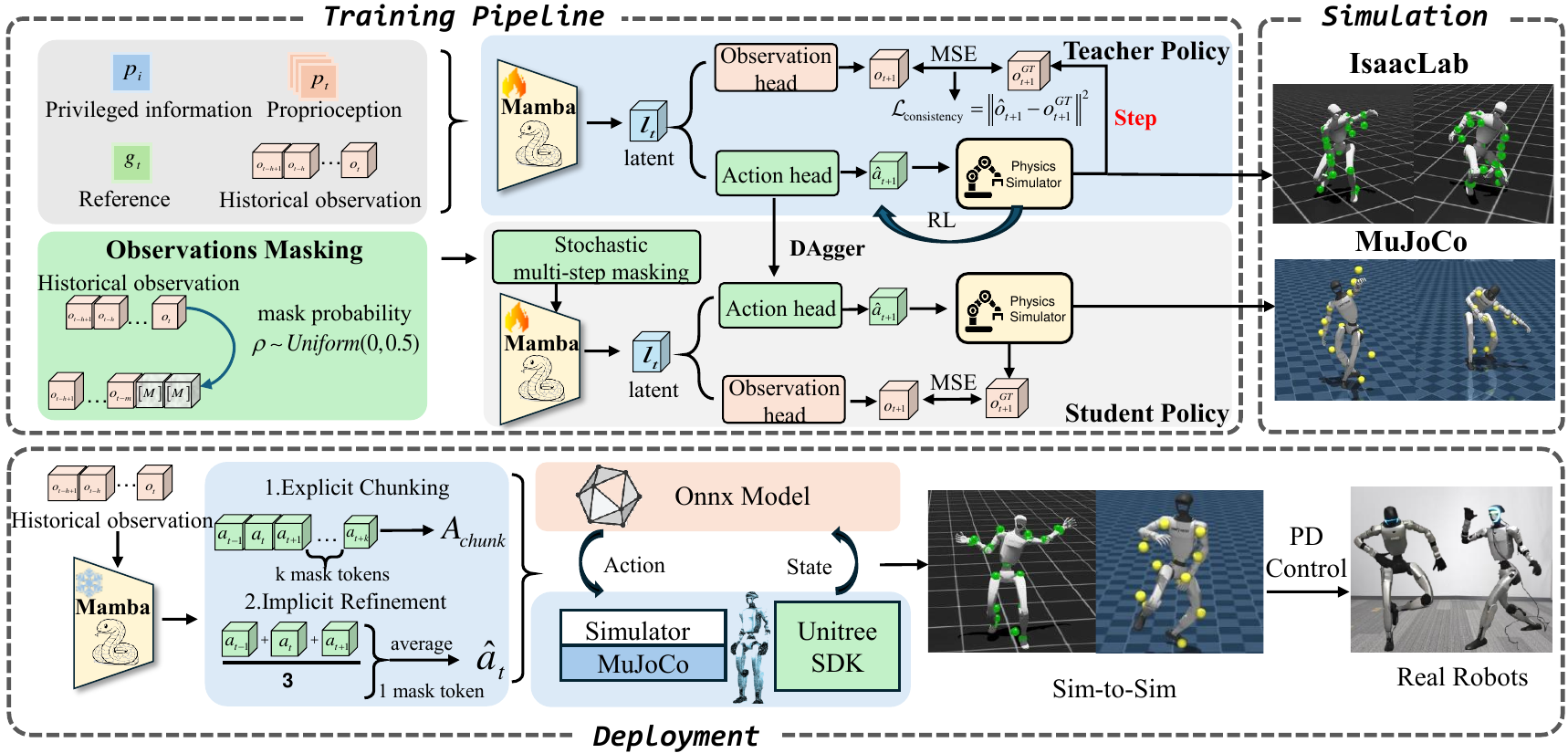}
    \caption{Overview of RoboDreamer. Our framework features a Mamba-based training pipeline that uses privileged information and longer historical observations to train a teacher policy via RL, while a student policy is trained with DAgger under stochastic multi-step masking. The temporal backbone enables explicit action chunking and implicit refinement through masked multi-step extrapolation.}
    \label{fig:framework}
\end{figure*}
\section{Method}
\label{sec:method}

This section details the architecture and training pipeline of the RoboDreamer, which is depicted in Figure~\ref{fig:framework}. We first give an overview in Section~\ref{subsec:overview}, introducing our physics-aligned temporal paradigm and its core rationale. Section~\ref{subsec:problem_formulation} formalizes humanoid locomotion as a partially observable Markov decision process and presents the mathematical foundation of the selective state-space model for our policy backbone. The learning process is split into two sequential stages: Section~\ref{subsec:stage1} elaborates on Phase I for predictive world modeling via mask prediction, and Section~\ref{subsec:stage2} on Phase II for anticipatory synthesis with stochastic multi-step masking. Finally, Section~\ref{subsec:inference} introduces our dual-mode inference paradigms, including explicit action chunking and implicit temporal refinement. Additional implementation details and hyperparameters are provided in the Appendix.

\subsection{Overview}
\label{subsec:overview}

RoboDreamer unifies perception, temporal reasoning, and actuation within a selective state-space backbone, shifting humanoid control from reactive frame-wise mapping to a temporally-aware paradigm. By leveraging Mamba's continuous-time formulation, the policy maintains an evolving latent state that aligns with underlying physical constraints.

Central to this robustness is Mamba's selective scan mechanism, which provides dynamic temporal filtering. It enables the model to mitigate sensory noise and handle visual occlusions by relying on its internal state, ensuring control continuity under severe sensory degradation. Beyond perception, we introduce a consistency loss for future-state prediction. This constraint encourages the latent space to capture the physical consequences of actions, allowing the agent to anticipate movement outcomes before execution.

Finally, the temporal extrapolation learned through masking naturally supports action chunking. The policy can generate multi-step future actions in a single forward pass, decoupling high-frequency motor execution from lower-frequency perception when desired.\subsection{Problem Formulation and State-Space Foundation}
\label{subsec:problem_formulation}

We formulate humanoid locomotion as a partially observable Markov decision process, where the agent must map a history of proprioceptive observations $\mathcal{O} = \{o_{t-h+1}, \dots, o_t\}$ to an optimal action $a_t$, where $h$ is history length. Traditional architectures often fail to capture the underlying continuous-time dynamics of humanoid systems. To address this, we employ the selective state-space model Mamba as the policy backbone.

The core of Mamba lies in its selective scan mechanism, governed by the following continuous-time linear system:
\begin{equation}
    h'(t) = \mathbf{A}h(t) + \mathbf{B}x(t), \quad y(t) = \mathbf{C}h(t).
\end{equation}
Through a data-dependent discretization step $\Delta = \text{Softplus}(\text{Parameter} + \text{Proj}(x_t))$, the system is transformed into a discrete recurrence:
\begin{equation}
    \overline{\mathbf{A}} = \exp(\Delta \mathbf{A}), \quad
    \overline{\mathbf{B}} =
    (\Delta \mathbf{A})^{-1}(\exp(\Delta \mathbf{A}) - \mathbf{I}) \cdot \Delta \mathbf{B}.
\end{equation}
This formulation allows RoboDreamer to maintain a latent state $h_t$ that evolves with the robot's dynamics. In our reported implementation, the full history window is processed at each control step, giving $O(L)$ sequence complexity for history length $L$.

\subsection{Stage I: Teacher Policy Training with Long-Horizon Context}
\label{subsec:stage1}

In the initial phase, we establish a high-fidelity motor prior by training a Mamba-based tracker policy that leverages an extended temporal window of historical observations. Unlike conventional MLP-based policies~\citep{chen2025gmt,li2025bfm,he2025asap,li2025language} that often use shorter temporal context, our approach incorporates a proprioceptive sequence $\mathcal{O}_{1:t} = \{o_{t-h+1}, \dots, o_t\}$. At 50 Hz, our default $h=35$ corresponds to 0.7 s of short-term temporal context, which captures local momentum, inertial evolution, and contact transitions.

To endow this long-memory network with a profound physical intuition, we employ a dual-head architecture: an action head optimized via Proximal Policy Optimization (PPO)~\citep{schulman2017proximal}, and an auxiliary observation head that predicts the subsequent state $\hat{o}_{t+1}$ based on the evolved latent state $h_t$. By minimizing the state consistency loss:
\begin{equation}
    \mathcal{L}_{cons}
    =
    \mathbb{E}_{h_t \sim \mathcal{P}}
    \left[
    \left\|
    \Psi(h_t) - o_{t+1}^{GT}
    \right\|^2
    \right].
\end{equation}
By distilling temporal information into a causal physical representation, this phase trains an oracle teacher policy $\pi_T$. Given its access to noise-free states over an extended temporal window, $\pi_T$ accurately characterizes the target locomotion manifold. This allows it to provide dense and high-quality supervision for the subsequent distillation process.

\subsection{Stage II: Student Policy Training via Masked Prediction}
\label{subsec:stage2}

The second phase elevates the agent from a reactive observer to an anticipatory intelligent system capable of surviving sensory blackouts and generating action chunks. In this stage, we transition from pure reinforcement learning to a teacher-student distillation framework. This transition is strategically designed to bypass the gradient instability and catastrophic divergence often encountered when introducing extreme sensory noise in high-dimensional reinforcement learning~\citep{li2017deep}.

\paragraph{Observations Masking}
To cultivate a policy that transcends reactive mapping and achieves true temporal autonomy, we introduce a randomized observations masking scheme. Unlike static noise injection, this strategy simulates severe sensory dropouts by intentionally blinding the agent to its most recent physical states. Specifically, we sample an activation probability $\rho \sim \text{Uniform}(0, c)$ (here $c$ is 0.5); for each activated environment, we randomly select a window length $m \in \{1, \dots, 5\}$ and mask the trailing sequence of observations $\{o_{t-m+1}, \dots, o_t\}$ with learnable $[\text{M}]$ tokens.

This deliberate sensory deprivation serves two strategic objectives. First, it compels the Mamba backbone to achieve predictive resilience, requiring the latent state $h(t)$ to transition from a reactive listener to a proactive predictor that can synthesize accurate actions $a_t$ solely by inpainting the missing reality from historical memories. Second, and more importantly, this masking regime serves as the essential prerequisite for generative action chunking. By learning to maintain stable locomotion during blind intervals, the policy effectively masters the capability of temporal extrapolation. This ensures that during inference, the model can treat future timestamps as masking tokens and successfully imagine a sequence of stable, executable actions in a single forward pass, effectively decoupling the frequency of motor actuation from the intermittent nature of sensory perception.

\paragraph{Supervised Distillation}
The policy from stage I acts as a frozen teacher $\pi_T$, providing stable, high-quality targets for the student policy $\pi_S$. The training objective is to minimize the discrepancy between the student's output and the teacher's action under full visibility:
\begin{equation}
    \mathcal{L}_{distill}
    =
    \mathbb{E}_{\mathcal{O} \sim \mathcal{D}}
    \left[
    \left\|
    \pi_S(\mathcal{O}_{masked}) - \pi_T(\mathcal{O}_{full})
    \right\|^2
    \right],
\end{equation}
By coupling this distillation loss with the continued consistency loss $\mathcal{L}_{cons}$, we compel the student to utilize Mamba's selection mechanism to ignore the $[\text{M}]$ tokens and instead rely on the internalized dynamics within its hidden state $h(t)$.

\subsection{Inference Paradigms}
\label{subsec:inference}

Following the dual-phase training, RoboDreamer internalizes the capacity to model historical context, anticipate future observations, and implicitly reason across multi-step action trajectories. We design two complementary inference strategies to fully exploit these capabilities: explicit action chunking generation and implicit action refinement. Both strategies reuse the temporal representation learned under masking; the latter enhances smoothness and stability by fusing adjacent predicted actions while retaining closed-loop replanning.

\paragraph{Explicit Action Chunking Generation}
This strategy facilitates multi-step action generation by appending learnable mask tokens $[\text{M}]$ to the observation sequence, directly applying the continuous masking resilience acquired during Phase II. Specifically, in the inference stage, we first construct the historical observation sequence $\mathcal{H}_t = \{o_{t-h+1}, \dots, o_t\}$ with a window length $h$. We then concatenate $k$ mask tokens to the end of this sequence, forming an augmented input sequence of total length $h+k$. A single forward pass through the trained Mamba policy yields the last $k+1$ action outputs from the action head:
\begin{equation}
    \mathbf{A}_{chunk}
    =
    \{a_t, a_{t+1}, \dots, a_{t+k}\},
\end{equation}
where $k$ is the action chunk size, a hyperparameter adjusted based on deployment requirements. These actions are then dispatched to the humanoid's low-level controller for sequential execution. This strategy provides an optional mode for reducing sensing frequency, but larger $k$ also increases the open-loop horizon and can reduce reactivity because later actions are executed without fresh feedback.

\paragraph{Implicit Action Refinement}
On the basis of action chunking, we design a more refined inference strategy to further improve the smoothness and stability of humanoid locomotion. This approach leverages Mamba's bidirectional temporal modeling to fuse historical and predicted future action information for current action refinement. This strategy is particularly effective when the action chunk size is set to three. We append a single mask token $[\text{M}]$ to the history $\mathcal{H}_t$, forming an input sequence of length $h+1$. After a single forward pass, we extract the last three outputs from the action head, corresponding to the triad $\{a_{t-1}, a_t, a_{t+1}\}$. Here, the policy automatically models the preceding action $a_{t-1}$ and anticipates the future action $a_{t+1}$ through its latent state evolution. We then compute the element-wise average to generate a refined current action $\bar{a}_t$:
\begin{equation}
    \bar{a}_t
    =
    \frac{1}{3}
    (a_{t-1} + a_t + a_{t+1})
\end{equation}

This refined strategy yields smoother and more stable locomotion than directly executing a longer open-loop chunk. The policy predicts adjacent actions $a_{t-1}$ and $a_{t+1}$, fuses them with the current action, executes only the refined current action, and replans at the next control step. This preserves closed-loop feedback while exploiting temporal predictions learned through masking.
\vspace{-4mm}
\section{Experiments}
We comprehensively evaluate the efficacy of our RoboDreamer on humanoid locomotion tasks, with a focus on validating the superiority of the Mamba-based policy over traditional architectures and the impact of key design choices, such as action chunk size, history length, and consistency loss. Additionally, we quantify computational complexity and provide qualitative visualizations to illustrate the stability advantage of RoboDreamer under sensory uncertainty and multi-step action generation scenarios. All experiments are conducted in IsaacLab~\cite{mittal2025isaac} and MuJoCo~\cite{todorov2012mujoco} physics simulators, with the Unitree G1 humanoid robot as the target platform for both simulation and real-world validation.

\begin{table*}[h]
\centering
\resizebox{\linewidth}{!}{
\footnotesize
\begin{tabular}{l|cccccccc}
\toprule
\multirow{2}{*}{Method} & \multicolumn{4}{c}{LAFAN1} & \multicolumn{4}{c}{FineDance} \\
\cmidrule(lr){2-5} \cmidrule(lr){6-9}
 & $E_{\text{mpkpe}} \downarrow$ & $E_{\text{mpjpe}} \downarrow$ & $E_{\text{vel}} \downarrow$ & $E_{\text{yaw-vel}} \downarrow$ & $E_{\text{mpkpe}} \downarrow$ & $E_{\text{mpjpe}} \downarrow$ & $E_{\text{vel}} \downarrow$ & $E_{\text{yaw-vel}} \downarrow$ \\
\midrule
\rowcolor{gray!20} 
\multicolumn{9}{c}{IsaacLab} \\
\midrule
Exbody2 & 62.47& 65.26 &0.373 & 0.443&
160.43 & 163.57 & 0.266 & 0.300\\
GMT & 47.32& 49.31& 0.326& 0.369&
155.81 & 157.92 & 0.254 & 0.287 \\
BeyondMimic &\cellcolor{orange!20}43.39&\cellcolor{red!15}44.07&0.287& 0.321&
154.70 & 156.04 & 0.251 & 0.284 \\
RoboPerform &44.60& 46.22 &\cellcolor{orange!20}0.284 & \cellcolor{orange!20}0.311&
\cellcolor{orange!20}152.29& \cellcolor{orange!20}154.58 &\cellcolor{orange!20}0.246 &\cellcolor{orange!20}0.278\\
Ours &\cellcolor{red!15}42.87& \cellcolor{orange!20}44.11 &\cellcolor{red!15}0.276 & \cellcolor{red!15}0.295&
\cellcolor{magenta!20}146.72 & \cellcolor{magenta!20}148.43 &\cellcolor{magenta!20}0.209 & \cellcolor{magenta!20}0.222 \\
\midrule
\midrule
\rowcolor{gray!20} 
\multicolumn{9}{c}{MuJoCo} \\
\midrule
Exbody2 & 81.64& 83.21 &0.366 & 0.449&
268.33 & 247.62 & 0.308 & 0.323 \\
GMT & 63.29& 65.31& 0.329& 0.364&
260.17 & 239.54 & 0.292 & 0.314 \\
BeyondMimic &\cellcolor{orange!20}58.47&\cellcolor{red!15}60.12&\cellcolor{orange!20}0.287 & \cellcolor{orange!20}0.314&
255.62 & 236.18 & 0.276 & 0.303 \\
RoboPerform &59.83& 61.55 &0.291& 0.327&
\cellcolor{orange!20}252.51& \cellcolor{orange!20}233.47 &\cellcolor{orange!20}0.250 & \cellcolor{orange!20}0.283\\
Ours &\cellcolor{red!15}57.11&\cellcolor{orange!20}58.76 &\cellcolor{red!15}0.279 & \cellcolor{red!15}0.298&
\cellcolor{magenta!20}246.33 & \cellcolor{magenta!20}227.92 &\cellcolor{magenta!20} 0.226 & \cellcolor{magenta!20}0.241 \\
\bottomrule
\end{tabular}
}
\caption{Quantitative comparison of tracking performance on FineDance and LAFAN1.}
\label{tab:tracking_performance}
\end{table*}

\begin{figure*}[h]
\centering
  \includegraphics[width=\textwidth]{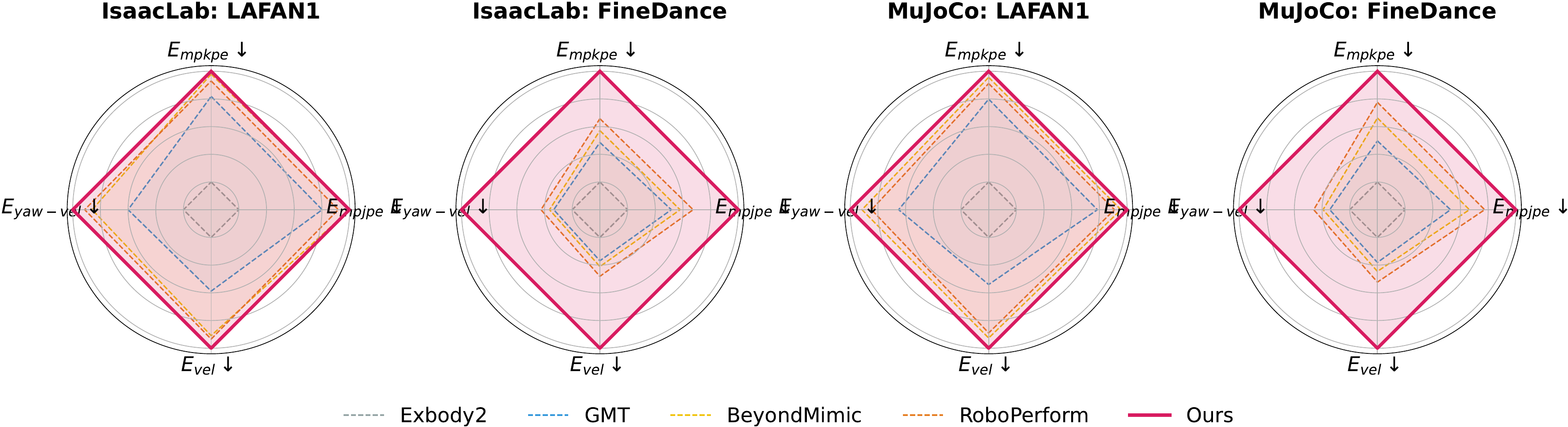}
\caption{Quantitative performance comparison across different datasets and simulators.}
\label{fig:radar}
\end{figure*}

\begin{table*}[t]
\centering
\footnotesize
\setlength{\tabcolsep}{3pt} 
\begin{tabular}{l|cccc|cc}
\toprule
\multirow{2}{*}{Method} & \multicolumn{4}{c}{FineDance} & \multirow{2}{*}{FLOPs $\downarrow$} & \multirow{2}{*}{Latency (ms) $\downarrow$} \\
\cmidrule(lr){2-5}
& $E_{\text{mpkpe}} \downarrow$ & $E_{\text{mpjpe}} \downarrow$ & $E_{\text{vel}} \downarrow$ & $E_{\text{yaw-vel}} \downarrow$ & & \\
\midrule
\rowcolor{gray!20} 
\multicolumn{7}{c}{IsaacLab} \\
\midrule
MLP-based & 163.35 & 165.26 & 0.278 & 0.324 & \cellcolor{magenta!20}6.9M & \cellcolor{magenta!20}3.5 \\
Transformer-based & \cellcolor{orange!20}149.79 & \cellcolor{orange!20}152.38 & \cellcolor{orange!20}0.217 & \cellcolor{orange!20}0.231 & 58.8M & 9.8 \\
Mamba-based & \cellcolor{magenta!20}146.72 & \cellcolor{magenta!20}148.43 & \cellcolor{magenta!20}0.209 & \cellcolor{magenta!20}0.222 & \cellcolor{orange!20}43.1M & \cellcolor{orange!20}4.4 \\
\bottomrule
\end{tabular}
\caption{Tracking performance and computational metrics (in the first stage) on a single environment of Mamba, MLP, and Transformer policies on FineDance.}
\label{tab:arch_compare}
\end{table*}

\begin{table*}[t]
\footnotesize
\setlength{\tabcolsep}{4pt}
\centering
\begin{minipage}[b]{0.6\linewidth}
\centering
\begin{tabular}{ccccc}
\toprule
\multirow{2}{*}{$k$} & \multicolumn{2}{c}{w/ consistency loss} & \multicolumn{2}{c}{w/o consistency loss} \\
\cmidrule(lr){2-3} \cmidrule(lr){4-5}
& $E_{\text{mpjpe}} \downarrow$ & $E_{\text{mpkpe}} \downarrow$
& $E_{\text{mpjpe}} \downarrow$ & $E_{\text{mpkpe}} \downarrow$ \\
\midrule
1 & \cellcolor{color1}42.87&\cellcolor{color2}44.11 & 43.93 & 45.26 \\
2 & 44.62 & 46.30 & \cellcolor{color1}48.01 & \cellcolor{color1}50.34 \\
3 & 78.72 & 81.29 & 85.72 & 88.09 \\
4 & 132.21 & 135.94 & 158.21 & 162.77 \\
Average & \cellcolor{color2}42.05 & \cellcolor{color1}44.33 & \cellcolor{color2}43.68 & \cellcolor{color2}45.24 \\
\bottomrule
\end{tabular}
\caption{Performance on different action chunk sizes and consistency loss in LAFAN1.}
\label{tab:chunk_size}
\end{minipage}
\hfill
\begin{minipage}[b]{0.35\linewidth}
\centering
\begin{tabular}{ccc}
\toprule
$h$ & $E_{\text{mpjpe}} \downarrow$ & $E_{\text{mpkpe}} \downarrow$ \\
\midrule
20 & 44.29 & 46.57  \\
25 & 44.32 & 46.44  \\
30 & 43.18 & \cellcolor{color1}44.93  \\
35 & \cellcolor{color2}42.87 & \cellcolor{color2}44.11 \\
40 & \cellcolor{color1}43.09& 45.12 \\
45 & 44.17 & 46.07 \\
\bottomrule
\end{tabular}
\caption{Performance across different history lengths.}
\label{tab:history_len}
\end{minipage}
\end{table*}

\subsection{Experimental Results}

\subsubsection{Tracking Performance}
To verify the effectiveness of RoboDreamer, we conduct performance comparisons with several state-of-the-art tracking policies. These methods include both teacher-student policies and pure tracker policies. We evaluate tracking performance on the AMASS test set, LAFAN, and FineDance datasets in both IsaacLab and MuJoCo simulation environments, respectively. We retrain all competing methods on our dataset with the same number of training iterations. For RoboPerform~\cite{li2025you}, we only test the teacher policy from its first training stage.
The results are presented in Table \ref{tab:tracking_performance} and Figure \ref{fig:radar}.

\subsubsection{Comparison with Traditional Architectures}
To further validate the superiority of the Mamba-based policy over MLP- and Transformer-based counterparts, we train MLP and Transformer-based policies with nearly equivalent parameter counts using the identical two-stage masked training strategy, and evaluate their tracking performance, computational complexity, and inference latency on the FineDance dataset, which is presented in Table \ref{tab:arch_compare}. We find that while the Mamba-based policy incurs slightly higher computational complexity, its inference latency is nearly on par with that of the MLP-based policy. More importantly, it achieves superior tracking performance.

\subsubsection{Ablation Studies}
We conduct relevant ablation studies on multiple factors, including the effects of different action chunk sizes $k$, consistency loss, and varying history lengths on the final results.
\paragraph{Action Chunk Size}
Since we perform consecutive masking on the current observation and previous observations in the second stage, the model is endowed with the ability to generate multiple future actions based on a sequence of masked tokens. 
 
We observe a performance trade-off regarding the chunk size: while a larger $k$ enhances temporal smoothness and reduces inference overhead, it introduces accumulated open-loop drift that exceeds the robot's physical stability margin. In addition, we tested an alternative approach to utilizing action chunks, which calculates the mean of historical and future actions as the current action. The results are presented in Table \ref{tab:chunk_size}.
\vspace{-2mm}
\paragraph{History Length}
Table \ref{tab:history_len} shows the effect of the historical observation window length $h = (20, 25, 30, 35, 40, 45)$. We observe that $h = 35$ achieves the best trade-off between tracking performance and computational cost.

\paragraph{Mask Probability}
To investigate what masking probability during training yields the best inference-time performance, we conduct an ablation study on the maximum masking probability $c$. Specifically, we train policies with masking ratios uniformly sampled from 
[0, 0.3], [0, 0.4], [0, 0.5], and [0, 0.6], respectively, and evaluate their tracking performance at inference time under the same masking conditions. As shown in Table \ref{mask}, the policy achieves the best inference performance when trained with masking ratios uniformly sampled from [0,0.5].

\begin{table*}[h]
\centering
\resizebox{\linewidth}{!}{
\footnotesize
\begin{tabular}{c|cccccccc}
\toprule
\multirow{2}{*}{$c$} & \multicolumn{4}{c}{LAFAN1} & \multicolumn{4}{c}{FineDance} \\
\cmidrule(lr){2-5} \cmidrule(lr){6-9}
 & $E_{\text{mpkpe}} \downarrow$ & $E_{\text{mpjpe}} \downarrow$ & $E_{\text{vel}} \downarrow$ & $E_{\text{yaw-vel}} \downarrow$ & $E_{\text{mpkpe}} \downarrow$ & $E_{\text{mpjpe}} \downarrow$ & $E_{\text{vel}} \downarrow$ & $E_{\text{yaw-vel}} \downarrow$ \\
\midrule
\rowcolor{gray!20} 
\multicolumn{9}{c}{IsaacLab} \\
\midrule
0.3 & 58.23 & 60.47 & 0.348 & 0.387 & 156.88 & 159.24 & 0.255 & 0.279 \\
0.4 & 52.31 & 54.78 & 0.335 & 0.358 & \cellcolor{orange!20}152.18 & \cellcolor{orange!20}154.06 & \cellcolor{orange!20}0.235 & \cellcolor{orange!20}0.256\\
0.5 &\cellcolor{red!15}42.87& \cellcolor{red!15}44.11 &\cellcolor{red!15}0.276 & \cellcolor{red!15}0.295&
\cellcolor{magenta!20}146.72 & \cellcolor{magenta!20}148.43 &\cellcolor{magenta!20}0.209 & \cellcolor{magenta!20}0.222 \\
0.6 & \cellcolor{orange!20}48.95 & \cellcolor{orange!20}50.72 & \cellcolor{orange!20}0.312 & \cellcolor{orange!20}0.334 & 155.43 & 157.89 & 0.248 & 0.271 \\
\bottomrule
\end{tabular}
}
\caption{Quantitative comparison of tracking performance on FineDance and LAFAN1.}
\label{mask}
\end{table*}

\paragraph{Consistency Loss}
Table \ref{tab:chunk_size} validates the necessity of consistency loss. We observe that the effect of consistency loss on tracking performance intensifies with increasing action chunk size. This is due to the consistency loss, empowering the policy to both focus on historical data and accurately model the future, which is an essential capability for generating future action chunks.


\section{Conclusion}
\label{sec:conclusion}

We present RoboDreamer, a predictive humanoid control framework centered on masked predictive distillation. The two-stage training combines next-observation consistency with randomized temporal masking, enabling the policy to infer missing recent information from history and to reuse the same masking interface for implicit closed-loop refinement and optional action chunking. Mamba serves as the temporal backbone, while our results indicate that masking/distillation drives a substantial portion of the gains and Mamba provides additional tracking improvements under matched settings. Our current real-world evaluation is limited to Unitree G1 without systematically injected sensing disturbances; broader corruption patterns, terrain changes, and cross-morphology transfer remain future work.


\bibliography{paper}

@String(TOG   = {ACM Trans. Graph.})

@String(TOG   = {ACM TOG})

@article{schulman2017proximal,
  title={Proximal policy optimization algorithms},
  author={Schulman, John and Wolski, Filip and Dhariwal, Prafulla and Radford, Alec and Klimov, Oleg},
  journal={arXiv preprint arXiv:1707.06347},
  year={2017}
}

@article{li2025language,
  title={From Language to Locomotion: Retargeting-free Humanoid Control via Motion Latent Guidance},
  author={Li, Zhe and Chi, Cheng and Wei, Yangyang and Zhu, Boan and Peng, Yibo and Huang, Tao and Wang, Pengwei and Wang, Zhongyuan and Zhang, Shanghang and Xu, Chang},
  journal={arXiv preprint arXiv:2510.14952},
  year={2025}
}

@article{he2025asap,
  title={Asap: Aligning simulation and real-world physics for learning agile humanoid whole-body skills},
  author={He, Tairan and Gao, Jiawei and Xiao, Wenli and Zhang, Yuanhang and Wang, Zi and Wang, Jiashun and Luo, Zhengyi and He, Guanqi and Sobanbab, Nikhil and Pan, Chaoyi and others},
  journal={arXiv preprint arXiv:2502.01143},
  year={2025}
}

@article{xie2025kungfubot,
  title={KungfuBot: Physics-Based Humanoid Whole-Body Control for Learning Highly-Dynamic Skills},
  author={Xie, Weiji and Han, Jinrui and Zheng, Jiakun and Li, Huanyu and Liu, Xinzhe and Shi, Jiyuan and Zhang, Weinan and Bai, Chenjia and Li, Xuelong},
  journal={arXiv preprint arXiv:2506.12851},
  year={2025}
}

@article{he2024omnih2o,
  title={Omnih2o: Universal and dexterous human-to-humanoid whole-body teleoperation and learning},
  author={He, Tairan and Luo, Zhengyi and He, Xialin and Xiao, Wenli and Zhang, Chong and Zhang, Weinan and Kitani, Kris and Liu, Changliu and Shi, Guanya},
  journal={arXiv preprint arXiv:2406.08858},
  year={2024}
}

@article{ji2024exbody2,
  title={Exbody2: Advanced expressive humanoid whole-body control},
  author={Ji, Mazeyu and Peng, Xuanbin and Liu, Fangchen and Li, Jialong and Yang, Ge and Cheng, Xuxin and Wang, Xiaolong},
  journal={arXiv preprint arXiv:2412.13196},
  year={2024}
}

@article{chen2025gmt,
  title={GMT: General Motion Tracking for Humanoid Whole-Body Control},
  author={Chen, Zixuan and Ji, Mazeyu and Cheng, Xuxin and Peng, Xuanbin and Peng, Xue Bin and Wang, Xiaolong},
  journal={arXiv preprint arXiv:2506.14770},
  year={2025}
}

@article{shao2025langwbc,
  title={LangWBC: Language-directed Humanoid Whole-Body Control via End-to-end Learning},
  author={Shao, Yiyang and Huang, Xiaoyu and Zhang, Bike and Liao, Qiayuan and Gao, Yuman and Chi, Yufeng and Li, Zhongyu and Shao, Sophia and Sreenath, Koushil},
  journal={arXiv preprint arXiv:2504.21738},
  year={2025}
}

@article{han2025kungfubot2,
  title={KungfuBot2: Learning Versatile Motion Skills for Humanoid Whole-Body Control},
  author={Han, Jinrui and Xie, Weiji and Zheng, Jiakun and Shi, Jiyuan and Zhang, Weinan and Xiao, Ting and Bai, Chenjia},
  journal={arXiv preprint arXiv:2509.16638},
  year={2025}
}

@article{peng2018deepmimic,
  title={Deepmimic: Example-guided deep reinforcement learning of physics-based character skills},
  author={Peng, Xue Bin and Abbeel, Pieter and Levine, Sergey and Van de Panne, Michiel},
  journal={ACM Transactions On Graphics (TOG)},
  volume={37},
  number={4},
  pages={1--14},
  year={2018},
  publisher={ACM New York, NY, USA}
}

@article{zhang2025hub,
  title={HuB: Learning Extreme Humanoid Balance},
  author={Zhang, Tong and Zheng, Boyuan and Nai, Ruiqian and Hu, Yingdong and Wang, Yen-Jen and Chen, Geng and Lin, Fanqi and Li, Jiongye and Hong, Chuye and Sreenath, Koushil and others},
  journal={arXiv preprint arXiv:2505.07294},
  year={2025}
}

@article{ze2025twist,
  title={Twist: Teleoperated whole-body imitation system},
  author={Ze, Yanjie and Chen, Zixuan and Ara{\'u}jo, Joao Pedro and Cao, Zi-ang and Peng, Xue Bin and Wu, Jiajun and Liu, C Karen},
  journal={arXiv preprint arXiv:2505.02833},
  year={2025}
}

@article{li2025clone,
  title={CLONE: Closed-Loop Whole-Body Humanoid Teleoperation for Long-Horizon Tasks},
  author={Li, Yixuan and Lin, Yutang and Cui, Jieming and Liu, Tengyu and Liang, Wei and Zhu, Yixin and Huang, Siyuan},
  journal={arXiv preprint arXiv:2506.08931},
  year={2025}
}

@article{wang2025experts,
  title={From experts to a generalist: Toward general whole-body control for humanoid robots},
  author={Wang, Yuxuan and Yang, Ming and Ding, Ziluo and Zhang, Yu and Zeng, Weishuai and Xu, Xinrun and Jiang, Haobin and Lu, Zongqing},
  journal={arXiv preprint arXiv:2506.12779},
  year={2025}
}

@article{liao2025beyondmimic,
  title={Beyondmimic: From motion tracking to versatile humanoid control via guided diffusion},
  author={Liao, Qiayuan and Truong, Takara E and Huang, Xiaoyu and Tevet, Guy and Sreenath, Koushil and Liu, C Karen},
  journal={arXiv preprint arXiv:2508.08241},
  year={2025}
}

@article{geyer2003positive,
  title={Positive force feedback in bouncing gaits?},
  author={Geyer, Hartmut and Seyfarth, Andre and Blickhan, Reinhard},
  journal={Proceedings of the Royal Society of London. Series B: Biological Sciences},
  volume={270},
  number={1529},
  pages={2173--2183},
  year={2003},
  publisher={The Royal Society}
}

@article{sreenath2011compliant,
  title={A compliant hybrid zero dynamics controller for stable, efficient and fast bipedal walking on MABEL},
  author={Sreenath, Koushil and Park, Hae-Won and Poulakakis, Ioannis and Grizzle, Jessy W},
  journal={The International Journal of Robotics Research},
  volume={30},
  number={9},
  pages={1170--1193},
  year={2011},
  publisher={SAGE Publications Sage UK: London, England}
}

@article{wang2025beamdojo,
  title={Beamdojo: Learning agile humanoid locomotion on sparse footholds},
  author={Wang, Huayi and Wang, Zirui and Ren, Junli and Ben, Qingwei and Huang, Tao and Zhang, Weinan and Pang, Jiangmiao},
  journal={arXiv preprint arXiv:2502.10363},
  year={2025}
}

@article{li2023robust,
  title={Robust and versatile bipedal jumping control through reinforcement learning},
  author={Li, Zhongyu and Peng, Xue Bin and Abbeel, Pieter and Levine, Sergey and Berseth, Glen and Sreenath, Koushil},
  journal={arXiv preprint arXiv:2302.09450},
  year={2023}
}

@article{huang2025learning,
  title={Learning humanoid standing-up control across diverse postures},
  author={Huang, Tao and Ren, Junli and Wang, Huayi and Wang, Zirui and Ben, Qingwei and Wen, Muning and Chen, Xiao and Li, Jianan and Pang, Jiangmiao},
  journal={arXiv preprint arXiv:2502.08378},
  year={2025}
}

@article{he2025learning,
  title={Learning getting-up policies for real-world humanoid robots},
  author={He, Xialin and Dong, Runpei and Chen, Zixuan and Gupta, Saurabh},
  journal={arXiv preprint arXiv:2502.12152},
  year={2025}
}

@inproceedings{li2025hold,
  title={Hold My Beer: Learning Gentle Humanoid Locomotion and End-Effector Stabilization Control},
  author={Li, Yitang and Zhang, Yuanhang and Xiao, Wenli and Pan, Chaoyi and Weng, Haoyang and He, Guanqi and He, Tairan and Shi, Guanya},
  booktitle={RSS 2025 Workshop on Whole-body Control and Bimanual Manipulation: Applications in Humanoids and Beyond}
}

@article{zhang2025falcon,
  title={FALCON: Learning Force-Adaptive Humanoid Loco-Manipulation},
  author={Zhang, Yuanhang and Yuan, Yifu and Gurunath, Prajwal and He, Tairan and Omidshafiei, Shayegan and Agha-mohammadi, Ali-akbar and Vazquez-Chanlatte, Marcell and Pedersen, Liam and Shi, Guanya},
  journal={arXiv preprint arXiv:2505.06776},
  year={2025}
}

@article{su2025hitter,
  title={Hitter: A humanoid table tennis robot via hierarchical planning and learning},
  author={Su, Zhi and Zhang, Bike and Rahmanian, Nima and Gao, Yuman and Liao, Qiayuan and Regan, Caitlin and Sreenath, Koushil and Sastry, S Shankar},
  journal={arXiv preprint arXiv:2508.21043},
  year={2025}
}

@article{peng2021amp,
  title={Amp: Adversarial motion priors for stylized physics-based character control},
  author={Peng, Xue Bin and Ma, Ze and Abbeel, Pieter and Levine, Sergey and Kanazawa, Angjoo},
  journal={ACM Transactions on Graphics (ToG)},
  volume={40},
  number={4},
  pages={1--20},
  year={2021},
  publisher={ACM New York, NY, USA}
}

@inproceedings{huang2010lcm,
  title={LCM: Lightweight communications and marshalling},
  author={Huang, Albert S and Olson, Edwin and Moore, David C},
  booktitle={2010 IEEE/RSJ International Conference on Intelligent Robots and Systems},
  pages={4057--4062},
  year={2010},
  organization={IEEE}
}

@article{gu2024humanoid,
  title={Humanoid-gym: Reinforcement learning for humanoid robot with zero-shot sim2real transfer},
  author={Gu, Xinyang and Wang, Yen-Jen and Chen, Jianyu},
  journal={arXiv preprint arXiv:2404.05695},
  year={2024}
}

@article{li2025you,
  title={Do you have freestyle? expressive humanoid locomotion via audio control},
  author={Li, Zhe and Chi, Cheng and Wei, Yangyang and Zhu, Boan and Huang, Tao and Sun, Zhenguo and Peng, Yibo and Wang, Pengwei and Wang, Zhongyuan and Liu, Fangzhou and others},
  journal={arXiv preprint arXiv:2512.23650},
  year={2025}
}

@article{li2025robomirror,
  title={Robomirror: Understand before you imitate for video to humanoid locomotion},
  author={Li, Zhe and Chi, Cheng and Zhu, Boan and Wei, Yangyang and Bai, Shuanghao and Ji, Yuheng and Peng, Yibo and Huang, Tao and Wang, Pengwei and Wang, Zhongyuan and others},
  journal={arXiv preprint arXiv:2512.23649},
  year={2025}
}

@article{vaswani2017attention,
  title={Attention is all you need},
  author={Vaswani, Ashish and Shazeer, Noam and Parmar, Niki and Uszkoreit, Jakob and Jones, Llion and Gomez, Aidan N and Kaiser, {\L}ukasz and Polosukhin, Illia},
  journal={Advances in neural information processing systems},
  volume={30},
  year={2017}
}

@inproceedings{gu2024mamba,
  title={Mamba: Linear-time sequence modeling with selective state spaces},
  author={Gu, Albert and Dao, Tri},
  booktitle={First conference on language modeling},
  year={2024}
}

@article{ma2026robust,
  title={Robust and Generalized Humanoid Motion Tracking},
  author={Ma, Yubiao and Yu, Han and Xie, Jiayin and Lv, Changtai and Luo, Qiang and Zhang, Chi and Yin, Yunpeng and Xing, Boyang and Ren, Xuemei and Zheng, Dongdong},
  journal={arXiv preprint arXiv:2601.23080},
  year={2026}
}

@article{harvey2020robust,
  title={Robust motion in-betweening},
  author={Harvey, F{\'e}lix G and Yurick, Mike and Nowrouzezahrai, Derek and Pal, Christopher},
  journal={ACM Transactions on Graphics (TOG)},
  volume={39},
  number={4},
  pages={60--1},
  year={2020},
  publisher={ACM New York, NY, USA}
}

@inproceedings{li2023finedance,
  title={Finedance: A fine-grained choreography dataset for 3d full body dance generation},
  author={Li, Ronghui and Zhao, Junfan and Zhang, Yachao and Su, Mingyang and Ren, Zeping and Zhang, Han and Tang, Yansong and Li, Xiu},
  booktitle={Proceedings of the IEEE/CVF International Conference on Computer Vision},
  pages={10234--10243},
  year={2023}
}

@inproceedings{peng2018sim,
  title={Sim-to-real transfer of robotic control with dynamics randomization},
  author={Peng, Xue Bin and Andrychowicz, Marcin and Zaremba, Wojciech and Abbeel, Pieter},
  booktitle={2018 IEEE international conference on robotics and automation (ICRA)},
  pages={3803--3810},
  year={2018},
  organization={IEEE}
}

@inproceedings{rudin2022learning,
  title={Learning to walk in minutes using massively parallel deep reinforcement learning},
  author={Rudin, Nikita and Hoeller, David and Reist, Philipp and Hutter, Marco},
  booktitle={Conference on robot learning},
  pages={91--100},
  year={2022},
  organization={PMLR}
}

@article{mittal2025isaac,
  title={Isaac lab: A gpu-accelerated simulation framework for multi-modal robot learning},
  author={Mittal, Mayank and Roth, Pascal and Tigue, James and Richard, Antoine and Zhang, Octi and Du, Peter and Serrano-Munoz, Antonio and Yao, Xinjie and Zurbr{\"u}gg, Ren{\'e} and Rudin, Nikita and others},
  journal={arXiv preprint arXiv:2511.04831},
  year={2025}
}

@article{li2026w1,
  title={FRoM-W1: Towards General Humanoid Whole-Body Control with Language Instructions},
  author={Li, Peng and Zhuang, Zihan and Gao, Yangfan and Dong, Yi and Li, Sixian and Jiang, Changhao and Dou, Shihan and Xi, Zhiheng and Zhou, Enyu and Huang, Jixuan and others},
  journal={arXiv preprint arXiv:2601.12799},
  year={2026}
}

@article{xie2026textop,
  title={TextOp: Real-time Interactive Text-Driven Humanoid Robot Motion Generation and Control},
  author={Xie, Weiji and Zheng, Jiakun and Han, Jinrui and Shi, Jiyuan and Zhang, Weinan and Bai, Chenjia and Li, Xuelong},
  journal={arXiv preprint arXiv:2602.07439},
  year={2026}
}

@article{li2025bfm,
  title={Bfm-zero: A promptable behavioral foundation model for humanoid control using unsupervised reinforcement learning},
  author={Li, Yitang and Luo, Zhengyi and Zhang, Tonghe and Dai, Cunxi and Kanervisto, Anssi and Tirinzoni, Andrea and Weng, Haoyang and Kitani, Kris and Guzek, Mateusz and Touati, Ahmed and others},
  journal={arXiv preprint arXiv:2511.04131},
  year={2025}
}

@article{luo2025sonic,
  title={Sonic: Supersizing motion tracking for natural humanoid whole-body control},
  author={Luo, Zhengyi and Yuan, Ye and Wang, Tingwu and Li, Chenran and Chen, Sirui and Castaneda, Fernando and Cao, Zi-Ang and Li, Jiefeng and Minor, David and Ben, Qingwei and others},
  journal={arXiv preprint arXiv:2511.07820},
  year={2025}
}

@article{li2017deep,
  title={Deep reinforcement learning: An overview},
  author={Li, Yuxi},
  journal={arXiv preprint arXiv:1701.07274},
  year={2017}
}

@inproceedings{todorov2012mujoco,
  title={MuJoCo: A physics engine for model-based control},
  author={Todorov, Emanuel and Erez, Tom and Tassa, Yuval},
  booktitle={2012 IEEE/RSJ International Conference on Intelligent Robots and Systems},
  pages={5026--5033},
  year={2012},
  organization={IEEE},
  doi={10.1109/IROS.2012.6386109}
}

@article{chung2014empirical,
  title={Empirical evaluation of gated recurrent neural networks on sequence modeling},
  author={Chung, Junyoung and Gulcehre, Caglar and Cho, KyungHyun and Bengio, Yoshua},
  journal={arXiv preprint arXiv:1412.3555},
  year={2014}
}

@article{hochreiter1997long,
  title={Long short-term memory},
  author={Hochreiter, Sepp and Schmidhuber, J{\"u}rgen},
  journal={Neural computation},
  volume={9},
  number={8},
  pages={1735--1780},
  year={1997},
  publisher={MIT press}
}

@inproceedings{wang2026humam,
  title={Humam: Humanoid motion control via end-to-end deep reinforcement learning with mamba},
  author={Wang, Yinuo and Tao, Xiaowen and Hu, Yunfeng and Liang, Yang and Song, Ziyu and Ding, Haitao and Zhou, Jinzhao},
  booktitle={2026 IEEE International Conference on Cybernetics and Intelligent Systems (CIS) and IEEE International Conference on Robotics, Automation and Mechatronics (RAM)},
  pages={319--324},
  year={2026},
  organization={IEEE}
}

@article{wang2026locomamba,
  title={LocoMamba: Vision-driven locomotion via end-to-end deep reinforcement learning with Mamba},
  author={Wang, Yinuo and Tao, Xiaowen},
  journal={Advanced Engineering Informatics},
  volume={70},
  pages={104230},
  year={2026},
  publisher={Elsevier}
}

@article{zhou2026humanoidmamba,
  title={Humanoidmamba: Generalized mamba-based policy learning with next-action prediction for humanoid locomotion},
  author={Zhou, Yi and Qiu, Jianbin and Zhang, Wei and Ni, Fenglei},
  journal={IEEE Transactions on Cognitive and Developmental Systems},
  year={2026},
  publisher={IEEE}
}

@misc{wang2026omnixtreme,
  title={OmniXtreme: Breaking the Generality Barrier in High-Dynamic Humanoid Control},
  author={Wang, Yunshen and Zhu, Shaohang and Zhi, Peiyuan and Li, Yuhan and Li, Jiaxin and Li, Yong-Lu and Xiao, Yuchen and Wang, Xingxing and Jia, Baoxiong and Huang, Siyuan},
  year={2026},
  eprint={2602.23843},
  archivePrefix={arXiv},
  primaryClass={cs.RO}
}

@inproceedings{ross2011dagger,
  title={A Reduction of Imitation Learning and Structured Prediction to No-Regret Online Learning},
  author={Ross, Stephane and Gordon, Geoffrey and Bagnell, Drew},
  booktitle={Proceedings of the Fourteenth International Conference on Artificial Intelligence and Statistics},
  pages={627--635},
  volume={15},
  series={Proceedings of Machine Learning Research},
  year={2011},
  publisher={PMLR}
}

\clearpage
\appendix
\section*{Appendix}
\addcontentsline{toc}{section}{Appendix Overview}
This supplementary document provides extended technical specifications, hyperparameter configurations, and additional experimental results for RoboDreamer. The appendix is organized as follows:
\begin{itemize}
    \item \textbf{Section~\ref{sec:related}}: Related works, including humanoid whole-body control and humanoid motion tracking.
    \item    \textbf{Section~\ref{app:implement}}: Detailed elaboration on the training pipeline, including state representations, motion filtering, domain randomization, and reward formulations.
    \item \textbf{Section~\ref{add_exp}}: Extended ablation studies on Mamba architecture depth and the action chunking mechanism.
    \item \textbf{Section~\ref{qual}}: Qualitative visualizations across simulation environments and real-world humanoid platforms.
\end{itemize}

\section{Related Work}
\label{sec:related}

\subsection{Humanoid Whole-body Control}
\label{subsec:whole_body_control}

Classical model-based frameworks for whole-body control achieve high-precision execution by leveraging analytical dynamics and rigorous contact formulations; however, they are often hindered by the immense effort required for system identification and their fragility when encountering unmodeled physical phenomena~\citep{geyer2003positive,sreenath2011compliant}. While data-driven reinforcement learning (RL) has successfully mitigated these modeling burdens, which demonstrates agility in specialized tasks such as extreme terrain navigation and dynamic recovery~\citep{wang2025beamdojo,peng2021amp,li2023robust,huang2025learning,he2025learning}, these methods typically necessitate meticulous, task-specific reward tuning and often struggle to exhibit the fluid coordination characteristic of biological systems.

To manage the high dimensionality of humanoid control, some researchers have proposed bifurcating upper- and lower-body objectives into decoupled policies~\citep{zhang2025falcon,li2025hold}, though this partitioning may compromise global inter-limb synchronization. Others have explored hierarchical architectures to sequence complex maneuvers~\citep{su2025hitter}, which improves modularity but frequently introduces significant design overhead and computational bottlenecks.

In contrast, whole-body motion tracking offers a paradigm shift by adopting human demonstrations as a direct, dense supervisory signal~\citep{han2025kungfubot2}. This approach effectively bypasses the requirement for handcrafted rewards while fostering the emergence of globally synchronized and expressive behaviors across diverse repertoires. By unifying disparate motor skills under a single tracking objective, this paradigm provides a scalable and principled trajectory toward synthesizing human-like humanoid intelligence.

\subsection{Humanoid Motion Tracking}
\label{subsec:humanoid_motion_tracking}

Pursuit of lifelike maneuvers via human demonstration has transitioned from isolated clip imitation to mastery of universal skill repertoires. Early milestones like DeepMimic~\citep{peng2018deepmimic} established imitation foundations through phase-based tracking and randomized initializations. To bridge the sim-to-real gap for agile skills, ASAP~\citep{he2025asap} introduced a multi-stage delta-action framework, while specialized systems such as HuB~\citep{zhang2025hub} and KungfuBot~\citep{xie2025kungfubot} attained high-fidelity reproduction of aggressive, high-dynamic motions through meticulous motion conditioning.

The evolution toward consolidated controllers was catalyzed by OmniH2O~\citep{he2024omnih2o}, which demonstrated a universal policy capable of spanning diverse motion libraries. Subsequent advancements, including ExBody2~\citep{ji2024exbody2}, refined gait expressiveness through decomposed tracking targets, while TWIST~\citep{ze2025twist} and CLONE~\citep{li2025clone} attained precision in teleoperation settings for lower-dynamic tasks. More complex strategies have also emerged, such as BumbleBee's~\citep{wang2025experts} expert-distillation pipeline and GMT's~\citep{chen2025gmt} prioritized root-velocity tracking. BeyondMimic~\citep{liao2025beyondmimic} achieves high-fidelity single-motion tracking through crafted objectives and precise system identification, which are then distilled into a unified diffusion policy for task-specific control. KungfuBot2~\citep{han2025kungfubot2} introduces an Mixture-of-Experts framework to enable a general motion tracking policy that generalizes across a diverse set of skills. BFM-Zero~\citep{li2025bfm} proposes that off-policy unsupervised RL is a viable approach to train a behavioral foundation model for whole-body control of a humanoid robot. Sonic~\citep{luo2025sonic} achieves strong tracking performance but demands large-scale datasets and substantial computational resources. Besides, several modality-driven locomotion methods~\citep{li2025robomirror,li2026w1,xie2026textop}, including LangWBC~\citep{shao2025langwbc}, RoboGhost~\citep{li2025language}, and RoboPerform~\citep{li2025you}, rely on MLP-based tracker policies. Additionally, Sonic~\citep{luo2025sonic} achieves strong whole-body control performance by training on an extremely large-scale dataset. However, it is also based on an MLP architecture. Recently, RGMT~\citep{ma2026robust} utilizes Transformer~\citep{vaswani2017attention} as the backbone of the tracker policy, which can fuse extended temporal contexts. However, its self-attention mechanism incurs a quadratic computational cost that escalates rapidly with history length, hindering real-time deployment on high-frequency humanoid systems.

In contrast, state-space models like Mamba~\citep{gu2024mamba} offer a principled alternative, providing $O(L)$ scaling and $O(1)$ inference efficiency without sacrificing long-range temporal reasoning. Our framework, RoboDreamer, departs from fragile kinematic mimicry toward visually grounded performance control, enabling humanoids to function as semantically aligned and physically resilient imitators.

\section{Implementation Details}
\label{app:implement}
Our Mamba policy uses a 2-layer selective state-space backbone with embedding dimension 256, state dimension $d_{\text{state}}=16$, historical observation window length $h=35$. The reinforcement learning training hyperparameters are set as follows: discount factor $\gamma=0.99$, clip parameter $\epsilon=0.2$, entropy coefficient $0.01$, and consistency loss weight $\lambda=1.0$. The model is trained on a single NVIDIA GeForce RTX 4090 GPU. All baseline methods are trained under the identical optimization configuration and computational resources to ensure a fair comparison. 
\paragraph{Dataset}
We conduct evaluations across both physics simulation and real-world deployment settings. For simulation experiments, RoboDreamer is trained on a filtered LAFAN1~\cite{harvey2020robust} and FineDance~\cite{li2023finedance} datasets, incorporating domain randomization~\cite{peng2018sim} and action delay~\cite{rudin2022learning} to enhance generalization; evaluation is performed on the filtered LAFAN1 dataset and FineDance. All simulation training and evaluation are conducted in IsaacLab with 4096 parallel environments for computational efficiency.

\paragraph{Baselines}
To validate the effectiveness of RoboDreamer, we conduct comparative evaluations against state-of-the-art tracker policies, including Exbody2~\cite{ji2024exbody2}, GMT~\cite{chen2025gmt}, BeyondMimic~\cite{liao2025beyondmimic}, and RoboPerform~\cite{li2025you}. Furthermore, to further demonstrate the superiority and robustness of Mamba over traditional policy architectures, we train MLP and Transformer models with nearly equivalent parameter counts under our identical training setup for a fair comparison. Specifically, our Mamba backbone is configured with 2 layers, a state dimension $d_{\text{state}}=256$ and a model dimension $d_{\text{model}}=256$; the MLP is designed with 4 layers and a hidden width of 512; the Transformer is implemented with 2 layers, a model dimension $d_{\text{model}}=256$ and 4 attention heads.

\paragraph{Metrics}
We adopt a comprehensive set of evaluation metrics to assess the performance of RoboDreamer:
\begin{itemize}
    \item Mean Per Joint Position Error ($E_{\text{mpjpe}}$, mm) measures joint-level tracking accuracy by computing the average error in degrees of freedom (DoF) rotations between the reference and predicted motion.
    \item Mean Per Keybody Position Error ($E_{\text{mpkpe}}$, mm) assesses keypoint tracking performance based on the average positional discrepancy between reference and predicted keypoint trajectories.   
    \item Linear velocity error ($E_{\text{vel}}$, m/s) quantifies the accuracy of root linear velocity tracking by calculating the average Euclidean distance between the reference and predicted root linear velocity vectors in the horizontal plane.
    \item Yaw velocity error ($E_{\text{yaw-vel}}$, rad/s) evaluates the precision of rotational motion tracking by computing the average absolute error between the reference and predicted root yaw angular velocity values.      
\end{itemize}
    
In addition, we conduct a comparative evaluation of computational complexity between RoboDreamer and the MLP/Transformer baselines under identical experimental settings, where complexity is quantified by floating-point operations (FLOPs) per forward pass and inference latency (ms) to assess and compare the efficiency of each architecture on a single NVIDIA GeForce RTX 4090 GPU.

\paragraph{Training Details}
This section elaborates on the state representation employed for policy training, encompassing proprioceptive states, privileged information, and network hyperparameters. As outlined in Table \ref{propri}, the proprioceptive state components are consistent across the teacher and student policies, with one pivotal distinction: the student policy lacks the priviledged information.

Our proprioceptive states comprise joint positions, joint velocities, root angular velocity, root projected gravity, and corresponding data from 35 historical frames, as detailed in Table \ref{propri}. Privileged information, in conjunction with proprioceptive states, constitutes the observations for the critic network. The student policy, on the other hand, also incorporates proprioceptive states from 35 historical observations. Comprehensive specifications of the target state are presented in Table \ref{target}. Both policies generate 29-dimensional target joint positions as outputs.

The teacher policy is optimized using PPO \cite{schulman2017proximal} alongside an auxiliary next-observation prediction loss. This objective endows the policy with an internal world model capable of perceiving future states, integrating privileged simulator information, reference motion targets, longer historical observations, and proprioceptive states as its high-dimensional input. These input modalities are concatenated and fed into a Mamba-based actor network with $d_{\text{width}} = 256$,  $d_{\text{state}} = 16$, $d_{\text{conv}} = 4$, and expand = 1. And for critic network, we still use MLP backbone.

The student policy is trained under a DAgger-like paradigm, operating without privileged information. At the same time, we still constrain the model with the consistency loss during the distillation process, enabling it to both generate actions accurately when multiple observations are masked and perceive future states. During training, we randomly mask the current observation as well as $k$ consecutive previous observations $\{o_{t-k+1}, ..., o_{t}\}$ with learnable mask token [\text{M}]. Detailed hyperparameters for both policies, including learning rates, batch sizes, and mamba network configurations, are provided in Table \ref{hyper}.
\begin{table*}[h]
\centering
\begin{tabular}{cc}
\begin{minipage}[t]{0.51\textwidth}
\centering
\begin{tabular}{lr}
\toprule
\multicolumn{2}{c}{\textbf{Proprioceptive States}}\\
\midrule
State Component & Dim. \\
\midrule
DoF position & 29 $\times$ (1+34) \\
DoF velocity & 29 $\times$ (1+34)\\
Last action & 29 $\times$ (1+34)\\
Root angular velocity & 3 $\times$ (1+34)\\
Reference DoF position & 29 $\times$ (1+34)\\
Reference DoF velocity & 29 $\times$ (1+34)\\
Reference anchor orientation & 6 $\times$ (1+34)\\
\midrule
Total dim & $154 \times 35$ \\
\midrule
\multicolumn{2}{c}{\textbf{Privileged Information}}\\
\midrule
Root linear velocity & 3 \\
Root angular velocity & 3 \\
DoF position & 29  \\
DoF velocity & 29 \\
Body position & 42\\
Body orientation & 84 \\
Last action & 29 \\
Reference DoF position & 29 \\
Reference DoF velocity & 29 \\
Reference anchor orientation & 6 \\
Reference anchor position & 3 \\
\midrule
Total dim & 286 \\
\bottomrule
\end{tabular}
\caption{Proprioceptive states and privileged information.}
\label{propri}
\end{minipage}
\begin{minipage}[t]{0.45\textwidth}
\centering
\begin{tabular}{lc}
\toprule
\multicolumn{2}{c}{\textbf{Teacher Policy}}\\
\midrule
State Component & Dim. \\
\midrule
DoF position & 29 $\times$ (1+34) \\
DoF velocity & 29 $\times$ (1+34)\\
Last action & 29 $\times$ (1+34)\\
Root angular velocity & 3 $\times$ (1+34)\\
Reference DoF position & 29 $\times$ (1+34)\\
Reference DoF velocity & 29 $\times$ (1+34)\\
Reference anchor orientation & 6 $\times$ (1+34)\\
\midrule
Total dim & 145 $\times$ 35 \\
\midrule
\multicolumn{2}{c}{\textbf{Student Policy}}\\
\midrule
DoF position & 29 $\times$ (1+34) \\
DoF velocity & 29 $\times$ (1+34)\\
Last action & 29 $\times$ (1+34)\\
Root angular velocity & 3 $\times$ (1+34)\\
Reference DoF position & 29 $\times$ (1+34)\\
Reference DoF velocity & 29 $\times$ (1+34)\\
Reference anchor orientation & 6 $\times$ (1+34)\\
\midrule
Total dim & 145 $\times$ 35  \\
\bottomrule
\end{tabular}
\caption{Reference information in the teacher and student policies.}
\label{target}
\end{minipage}
\end{tabular}
\end{table*}

\begin{table}[h]
\centering

\begin{tabular}{lc}
\toprule
\textbf{Hyperparameter} & \textbf{Value} \\
\midrule
\quad Optimizer & Adam \\
\quad $\beta_1, \beta_2$ & 0.9, 0.999 \\
\quad Initial Learning Rate & $1\times10^{-3}$ \\
\quad Batch Size & 4096 \\
\midrule
\multicolumn{2}{c}{\textbf{Teacher Policy}} \\
\quad GAE Discount factor ($\gamma$) & 0.99 \\
\quad GAE Decay factor ($\gamma$) & 0.95 \\
\quad Clip Parameter & 0.2 \\
\quad Entropy Coefficient & 0.01 \\
\quad Max Gradient Norm & 1 \\
\quad Learning Epochs & 5 \\
\quad Mini Batches & 4 \\
\quad Value Loss Coefficient & 1.0 \\
\quad Value MLP Size & [512, 256, 128] \\
\quad Mamba Layers & 2 \\
\quad Mamba Width Dimension & 256 \\
\quad Mamba State Dimension & 16 \\
\quad Mamba Conv Dimension & 4 \\
\quad Mamba Expand & 1 \\
\midrule
\multicolumn{2}{c}{\textbf{Student Policy}} \\
\quad Mamba Layers & 2 \\
\quad Mamba Width Dimension & 128 \\
\quad Mamba State Dimension & 16 \\
\quad Mamba Conv Dimension & 4 \\
\quad Mamba Expand & 1 \\
\bottomrule
\end{tabular}
\caption{Hyperparameters for teacher and student policy training.}
\label{hyper}
\end{table}

\paragraph{Motion Filter and Retargeting}
We quantify motion stability following \cite{xie2025kungfubot}, using the ground-projected distance between the center of mass and center of pressure with a preset stability threshold for each frame. Let $\bar{\mathbf{p}}^{\text{CoM}}_t = (p^{\text{CoM}}_{t,x}, p^{\text{CoM}}_{t,y})$ and $\bar{\mathbf{p}}^{\text{CoP}}_t = (p^{\text{CoP}}_{t,x}, p^{\text{CoP}}_{t,y})$ denote the 2D ground projections of CoM and CoP at frame $t$. We define $\Delta d_t = \|\bar{\mathbf{p}}^{\text{CoM}}_t - \bar{\mathbf{p}}^{\text{CoP}}_t\|_2$ as this key distance metric. A frame qualifies as stable when $\Delta d_t < \epsilon_{\text{stab}}$. Motion sequences are retained only if their initial and final frames satisfy stability requirements and the longest contiguous unstable segment comprises fewer than 100 frames.

\paragraph{Simulator}
Aligning with standard protocols in motion tracking policy research \cite{ji2024exbody2, he2025asap}, we implement a three-stage evaluation pipeline. We first conduct large-scale reinforcement learning training in IsaacLab, then perform zero-shot transfer to MuJoCo for cross-simulator generalization assessment, and finally execute physical deployment on the Unitree G1 humanoid platform to validate real-world performance.

\paragraph{Reference State Initialization}
Task initialization is pivotal for reinforcement learning training. We find that naive episode initialization at the onset of reference motions frequently causes policy failure, especially for complex motion sequences. This issue induces environment overfitting to simpler frames while disregarding the most challenging motion segments.

To mitigate this limitation, we adopt the Reference State Initialization framework \cite{peng2018deepmimic}. We uniformly sample time-phase variables within the range $[0,1]$ to randomize the starting point of the reference motion that the policy must track. The robot's state, encompassing root position, orientation, linear and angular velocities, and joint positions and velocities, is then initialized to match the reference motion's values at the sampled phase. This strategy enhances motion tracking performance, particularly for highly dynamic whole-body motions, by enabling the policy to learn diverse movement segments in parallel instead of being restricted to strict sequential learning.

\paragraph{Domain Randomization and Regularization}
To boost the robustness and generalization capability of the pretrained policy, we employ a comprehensive set of domain randomization techniques and regularization strategies. Detailed specifications of these methods are presented in Table \ref{tab:domain_randomization}.

\begin{table}[h]
\centering
\renewcommand{\arraystretch}{1.4}
\begin{tabular}{cc}
\textbf{Domain Randomization} & \textbf{Sampling Distribution} \\
\hline
\textit{Physical parameters} & \\
Static friction coefficients & $\mu_{\text{static}} \sim \mathcal{U}[0.3, 1.6]$ \\
Dynamic friction coefficients & $\mu_{\text{dynamic}} \sim \mathcal{U}[0.3, 1.2]$ \\
Restitution coefficient & $e_{\text{rest}} \sim \mathcal{U}[0, 0.5]$ \\
Default joint positions [rad] & $\Delta \theta_j^0 \sim \mathcal{U}[-0.01, 0.01]$ \\
Torso’s COM offset [m] & $\Delta x \sim \mathcal{U}[-0.025, 0.025]$, $\Delta y \sim \mathcal{U}[-0.05, 0.05]$, \\
& $\Delta z \sim \mathcal{U}[-0.05, 0.05]$ \\
\hline
\textit{Root velocity perturbations} & \\
Root linear vel [m/s] & $v_x \sim \mathcal{U}[-0.5, 0.5]$, $v_y \sim \mathcal{U}[-0.5, 0.5]$, $v_z \sim \mathcal{U}[-0.2, 0.2]$ \\
Push duration [s] & $\Delta t \sim \mathcal{U}[1.0, 3.0]$ \\
Root angular vel [rad/s] & $\omega_x$, $\omega_y \sim \mathcal{U}[-0.52, 0.52]$, $\omega_z \sim \mathcal{U}[-0.78, 0.78]$ \\
\hline
\end{tabular}
\caption{Domain randomization parameters and their sampling distributions.}
\label{tab:domain_randomization}
\end{table}

\paragraph{Motion Tracking Rewards}
We design the reward function as a weighted combination of task-alignment rewards and regularization penalties, with detailed specifications provided in Table \ref{tab:reward_terms}. Following~\cite{liao2025beyondmimic}, this formulation prioritizes both tracking performance and motion physical plausibility for the humanoid robot.

\begin{table}[h]
\centering
\scriptsize  
\setlength{\tabcolsep}{2pt}  
\renewcommand{\arraystretch}{1.1}  
\begin{tabular}{l c r}
\textbf{Reward Terms} & \textbf{Equation} & \textbf{Weight} \\
\hline
\textit{Task (Tracking)} & & \\
Body Position & 
$\exp\left( -\left( \frac{1}{|\mathcal{B}_{\text{target}}|} \sum_{b \in \mathcal{B}_{\text{target}}} \|\mathbf{p}_b^{\text{des}} - \mathbf{p}_b\|^2 \right) / 0.3^2 \right)$ 
& 1.0 \\
Body Orientation & 
$\exp\left( -\left( \frac{1}{|\mathcal{B}_{\text{target}}|} \sum_{b \in \mathcal{B}_{\text{target}}} \|\log(R_b^{\text{des}} R_b^\top)\|^2 \right) / 0.4^2 \right)$ 
& 1.0 \\
Body Linear velocity & 
$\exp\left( -\left( \frac{1}{|\mathcal{B}_{\text{target}}|} \sum_{b \in \mathcal{B}_{\text{target}}} \|\mathbf{v}_b^{\text{des}} - \mathbf{v}_b\|^2 \right) / 1.0^2 \right)$ 
& 1.0 \\
Body Angular velocity & 
$\exp\left( -\left( \frac{1}{|\mathcal{B}_{\text{target}}|} \sum_{b \in \mathcal{B}_{\text{target}}} \|\omega_b^{\text{des}} - \omega_b\|^2 \right) / 3.14^2 \right)$ 
& 1.0 \\
Anchor Position& 
$\exp\left( -\|\mathbf{p}_{\text{anchor}}^{\text{des}} - \mathbf{p}_{\text{anchor}}\|^2 / 0.3^2 \right)$ 
& 0.5 \\
Anchor Orientation& 
$\exp\left( -\|\log(R_{\text{anchor}}^{\text{des}} R_{\text{anchor}}^\top)\|^2 / 0.4^2 \right)$ 
& 0.5 \\
\hline
\textit{Regularization} & & \\
Action smoothness & 
$\|\mathbf{a}_t - \mathbf{a}_{t-1}\|^2$ 
& $-0.1$ \\
Joint position limit & 
$\sum_{j=1}^N \left[ \max(l_j - \theta_j, 0) + \max(\theta_j - u_j, 0) \right]$ 
& $-10.0$ \\
Undesired self-contacts & 
$\sum_{b \notin \mathcal{B}_{\text{ee}}} \mathbf{1}\left[ \|f_b^{\text{self}}\| > 1\,\text{N} \right]$ 
& $-0.1$ \\
\hline
\end{tabular}
\caption{Reward function breakdown for tracking tasks, including task terms and regularization terms with corresponding weights.}
\label{tab:reward_terms}
\end{table}

\paragraph{Sim-to-Sim Transfer}Consistent with findings in Humanoid-Gym \cite{gu2024humanoid}, MuJoCo offers more realistic dynamic simulations compared to IsaacLab and IsaacGym. Following standard evaluation protocols in motion tracking policy research \cite{ji2024exbody2}, we conduct reinforcement learning training in IsaacLab to leverage its superior computational efficiency. To assess policy robustness and generalization capability, we perform zero-shot transfer to the MuJoCo simulator. This sim-to-sim transfer functions as an intermediate validation step prior to physical humanoid robot deployment, verifying the real-world motion tracking effectiveness of our framework.

\paragraph{Sim-to-Real Deployment}Real-world experiments are carried out on a Unitree G1 humanoid robot equipped with an onboard Jetson Orin NX module for computation and communication. The control policy processes motion tracking targets to generate target joint positions, then transmits control commands to the robot’s low-level controller at 50Hz with a communication latency ranging from 18ms to 30ms. The low-level controller operates at 500Hz to ensure stable real-time actuation. Communication between the high-level policy and low-level interface is realized through LCM \cite{huang2010lcm}.

\section{Additional Experiments}
\label{add_exp}
\paragraph{Different Mamba layers}
We conduct an ablation study on the number of layers of the Mamba network, as presented in Table \ref{tab: layer}. Given the relatively large number of parameters in Mamba, a deeper network architecture yields better tracking performance while incurring increased latency and higher GPU memory consumption. By comprehensively considering various performance metrics, we ultimately adopt a two-layer Mamba network for our model.

\begin{table*}[t]
\centering
\setlength{\tabcolsep}{7pt}      
\renewcommand{\arraystretch}{1.05} 
\scriptsize
\begin{tabular}{c|ccccc}
\toprule
\multirow{2}{*}{Layer} & \multicolumn{5}{c}{LAFAN1}\\
\cline{2-6}
 & $E_{\text{mpkpe}} \downarrow$ & $E_{\text{mpjpe}} \downarrow$ & $E_{\text{vel}} \downarrow$ & $E_{\text{yaw-vel}} \downarrow$ & Latency $\downarrow$\\
\midrule
\rowcolor{gray!10}
\multicolumn{6}{c}{IsaacLab} \\
\midrule
1 & 44.46 & 45.59 & 0.293 & 0.328 & \cellcolor{red!15}4.2 \\
2 & 42.87 & 44.11 & \cellcolor{red!15}0.276 & \cellcolor{red!15}0.295 & \cellcolor{orange!20}4.4 \\
3 & \cellcolor{orange!20}42.67 & \cellcolor{orange!20}44.03 & 0.282 & 0.311 & 7.9 \\
4 & \cellcolor{red!15}42.08 & \cellcolor{red!15}43.87 & \cellcolor{orange!20}0.278 & \cellcolor{orange!20}0.300 & 10.2 \\
\bottomrule
\end{tabular}
\caption{Quantitative comparison of tracking performance on LAFAN1 across different Mamba layers.}
\label{tab: layer}
\end{table*}

\paragraph{Action Chunk Size}
The most distinctive advantage of RoboDreamer over alternative methods is its ability to output multiple action chunks and execute continuous steps with a single forward pass of the model. In the main text, we report the tracking performance across different action chunk sizes, which are obtained by consecutively masking $k$ observations during the distillation phase, where $k$ is a random value sampled from 1 to 3. We find that a larger number of consecutively masked observations during training enables the model to generate more action chunks. In Table \ref{tab:acs1} and \ref{tab:acs2}, we present the tracking performance of different action chunks when $k$ is randomly sampled $\in [1, 4]$ and $[1, 5]$ during distillation, which verifies that the model can output an increasing number of action chunks as we progressively expand the number of consecutively masked observations.

\paragraph{Comparison with other $O(1)$ methods}
To further demonstrate the effectiveness of the Mamba-based policy, we compare it against alternative sequence models that also achieve $O(1)$ inference complexity per time step, including GRU~\cite{chung2014empirical} and LSTM~\cite{hochreiter1997long}. All baseline policies are trained under the same two-stage distillation framework with approximately equal parameter counts and identical observation history length. The results are reported in Table \ref{o1}.
\begin{table*}[t]
\centering
\footnotesize
\setlength{\tabcolsep}{3pt} 
\begin{tabular}{l|cccc|cc}
\toprule
\multirow{2}{*}{Method} & \multicolumn{4}{c}{FineDance} & \multirow{2}{*}{FLOPs $\downarrow$} & \multirow{2}{*}{Latency (ms) $\downarrow$} \\
\cmidrule(lr){2-5}
& $E_{\text{mpkpe}} \downarrow$ & $E_{\text{mpjpe}} \downarrow$ & $E_{\text{vel}} \downarrow$ & $E_{\text{yaw-vel}} \downarrow$ & & \\
\midrule
\rowcolor{gray!20} 
\multicolumn{7}{c}{IsaacLab} \\
\midrule
GRU & 152.37 & 153.96 & 0.224 & 0.251 & \cellcolor{magenta!20}40.7M & \cellcolor{magenta!20}4.1 \\
LSTM & 148.39 & 151.66 & 0.219 & 0.248 & 42.3M & 4.4 \\
Mamba-based & \cellcolor{magenta!20}146.72 & \cellcolor{magenta!20}148.43 & \cellcolor{magenta!20}0.209 & \cellcolor{magenta!20}0.222 & 43.1M & 4.4 \\
\bottomrule
\end{tabular}
\caption{Tracking performance and computational metrics (in the first stage) on a single environment of Mamba, GRU, and LSTM on FineDance.}
\label{o1}
\end{table*}

\paragraph{Action Chunking on Baseline Policies}
We also evaluate action chunking on the Transformer-based and MLP-based policies. In this experiment, the policy is configured to output three consecutive actions, which are then averaged element-wise to produce the final actuation command. The corresponding tracking performance is reported in Table \ref{ac_baselines}. We observe that when training MLP-based and Transformer-based policies under the same masking strategy, their performance degrades significantly compared to the Mamba-based policy. In most cases, the robot fails to maintain balance and falls.
\begin{table*}[t]
\centering
\footnotesize
\setlength{\tabcolsep}{3pt} 
\begin{tabular}{l|cccc}
\toprule
\multirow{2}{*}{Method} & \multicolumn{4}{c}{LAFAN1} \\
\cmidrule(lr){2-5}
& $E_{\text{mpkpe}} \downarrow$ & $E_{\text{mpjpe}} \downarrow$ & $E_{\text{vel}} \downarrow$ & $E_{\text{yaw-vel}} \downarrow$ \\
\midrule
\rowcolor{gray!20} 
\multicolumn{5}{c}{IsaacLab} \\
\midrule
MLP-based & 68.32 & 71.97 & 0.534 & 0.566\\
Transformer-based & 53.20 & 56.55 & 0.397 & 0.425\\
Mamba-based & \cellcolor{magenta!20}42.05 & \cellcolor{magenta!20}44.33 & \cellcolor{magenta!20}0.217 & \cellcolor{magenta!20}0.228 \\
\bottomrule
\end{tabular}
\caption{Tracking performance of action chunking ($k$=3 with temporal averaging) on MLP-based and Transformer-based policies on LAFAN1. Results are reported on a single environment.}
\label{ac_baselines}
\end{table*}

\begin{table}[t]
\footnotesize
\setlength{\tabcolsep}{3.5pt}
\centering

\begin{minipage}{0.48\linewidth}
\centering
\begin{tabular}{ccccc}
\toprule
\multirow{2}{*}{$k$} & \multicolumn{4}{c}{$k$ randomly sampled $\in [1, 4]$} \\
\cmidrule(lr){2-5}
& $E_{\text{mpjpe}} \downarrow$ & $E_{\text{mpkpe}} \downarrow$ & $E_{\text{vel}} \downarrow$ & $E_{\text{yaw-vel}} \downarrow$ \\
\midrule
1 & \cellcolor{color2}42.94&\cellcolor{color2}44.11 &\cellcolor{color2}0.278 &\cellcolor{color2}0.295 \\
2 & \cellcolor{color1}43.26&\cellcolor{color1}44.71 &\cellcolor{color1}0.285&\cellcolor{color1}0.301\\
3 & 56.93 & 59.20 & 0.327 & 0.341 \\
4 & 66.42 & 68.51 & 0.354 & 0.373\\
5 & 87.39 & 92.48 & 0.400 & 0.426 \\
\bottomrule
\end{tabular}
\caption{Performance on different action chunk sizes in LAFAN1 ($k \in [1,4]$).}
\label{tab:acs1}
\end{minipage}
\hfill
\begin{minipage}{0.48\linewidth}
\centering
\begin{tabular}{ccccc}
\toprule
\multirow{2}{*}{$k$} & \multicolumn{4}{c}{$k$ randomly sampled $\in [1, 5]$} \\
\cmidrule(lr){2-5}
& $E_{\text{mpjpe}} \downarrow$ & $E_{\text{mpkpe}} \downarrow$ & $E_{\text{vel}} \downarrow$ & $E_{\text{yaw-vel}} \downarrow$ \\
\midrule
1 & \cellcolor{color2}43.02&\cellcolor{color2}44.28 &\cellcolor{color2}0.281&\cellcolor{color2}0.299 \\
2 & \cellcolor{color1}43.42&\cellcolor{color1}44.78 &\cellcolor{color1}0.291&\cellcolor{color1}0.304\\
3 & 50.27 & 53.10 & 0.308 & 0.329 \\
4 & 57.32 & 60.06 & 0.333 & 0.350\\
5 & 65.87 & 67.34 & 0.384& 0.398 \\
\bottomrule
\end{tabular}
\caption{Performance on different action chunk sizes in LAFAN1 ($k \in [1,5]$).}
\label{tab:acs2}
\end{minipage}
\end{table}

\section{Qualitative Analysis and Real-world Deployment}
\label{qual}

\begin{figure*}[t]
\centering
  \includegraphics[width=\textwidth]{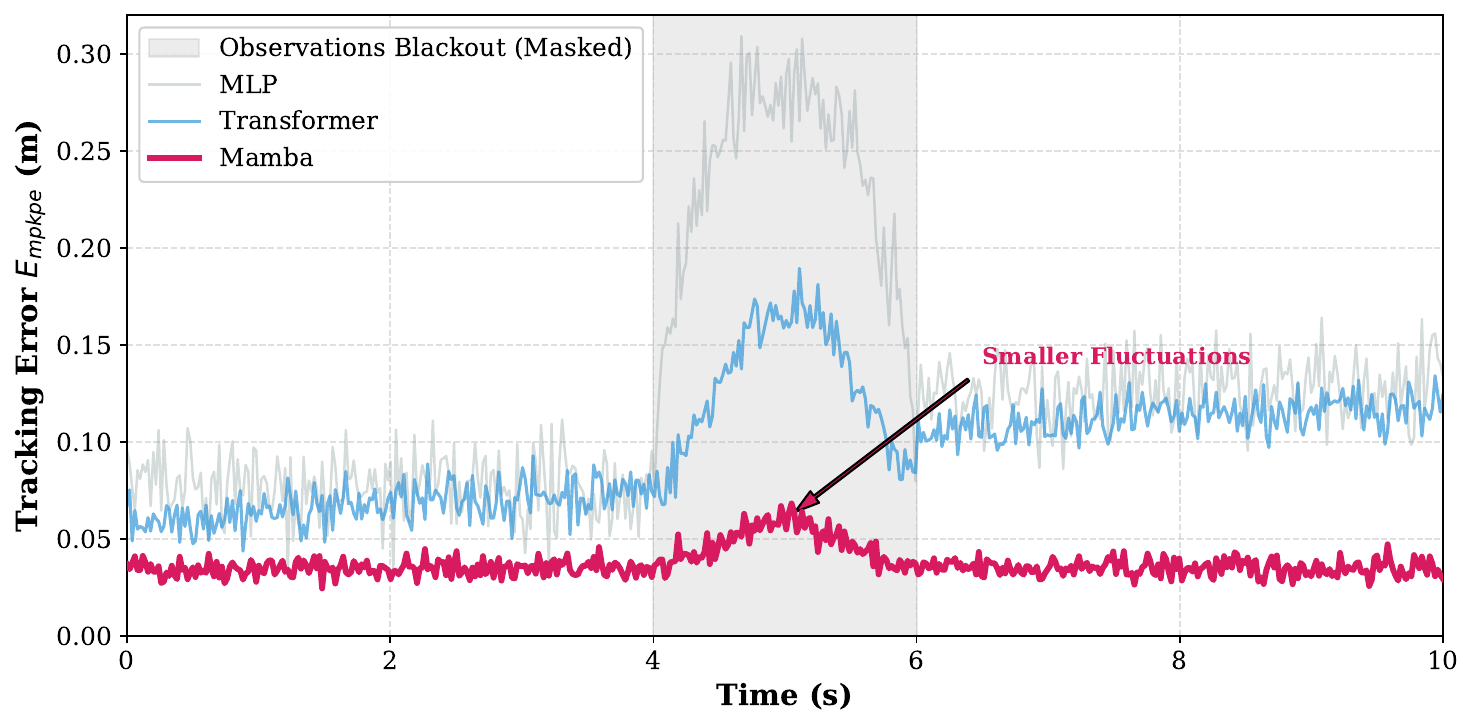}
\caption{Tracking error $E_{\text{mpkpe}}$ of different policies on the Charleston dance under observation masking.}
\label{fig:fluctuations}
\end{figure*}

\paragraph{Locomotion Stability}
To validate that RoboDreamer exhibits superior anti-interference capability and robustness when observations are masked, we quantify the tracking error $E_{\text{mpkpe}}$ (m) throughout the motion sequence with current observations masked. 
As illustrated in Figure \ref{fig:fluctuations}, the Mamba-based policy inherently yields a lower tracking error; furthermore, it demonstrates significantly stronger resilience to interference upon abrupt observation masking. In contrast, the MLP/Transformer-based policies both suffer from a substantial surge in tracking error under the same setting.

\begin{figure*}[t]
\centering
  \includegraphics[width=\textwidth]{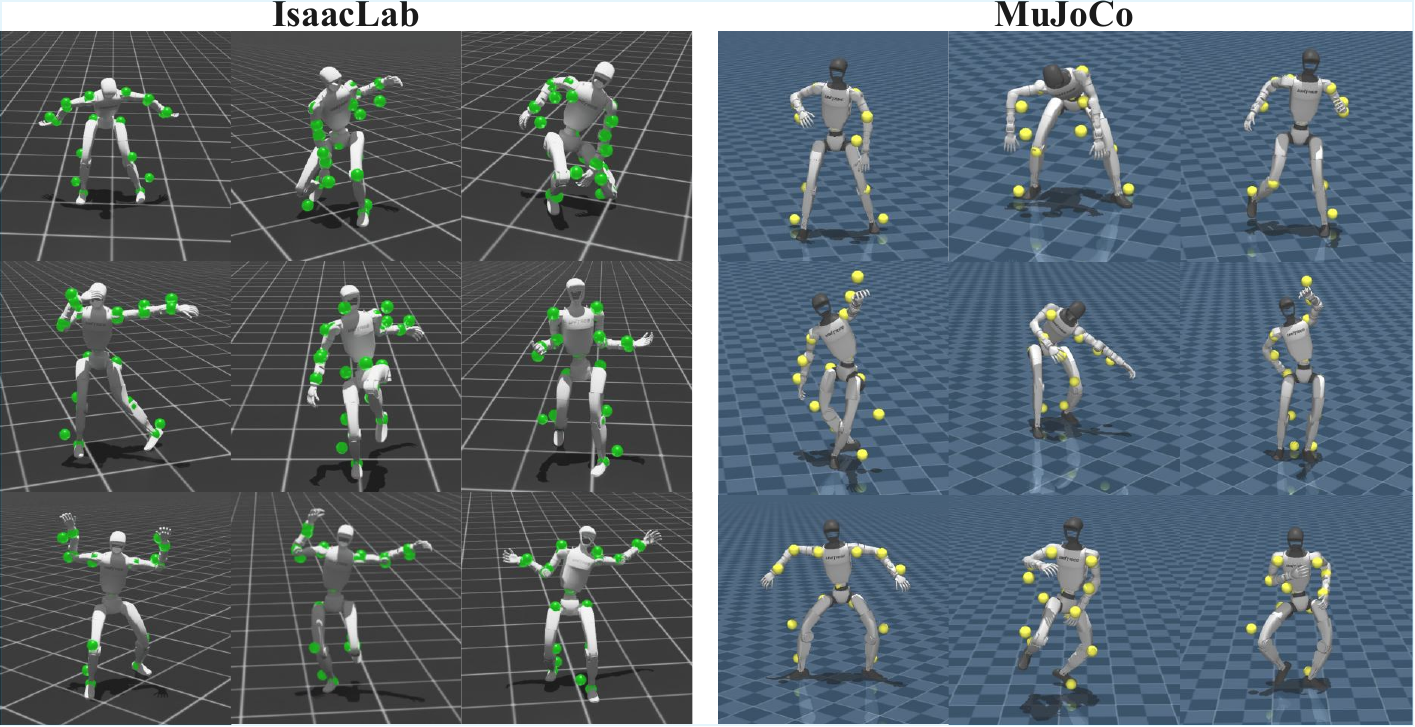}
\caption{Quantitative performance comparison across different datasets and simulators.}
\label{fig:sim}
\end{figure*}

\paragraph{Visualizations in Simulators}
We perform a qualitative evaluation of the motion tracking policy across three deployment scenarios: simulation (IsaacLab), cross-simulator transfer (MuJoCo), and real-world deployment on Unitree G1. Figure~\ref{fig:sim} showcases representative tracking sequences, demonstrating the policy’s capacity to retain balance during dynamic motion transitions and generalize effectively across distinct physics engines and hardware platforms. 

\begin{figure*}[t]
\centering
  \includegraphics[width=1.0\columnwidth]{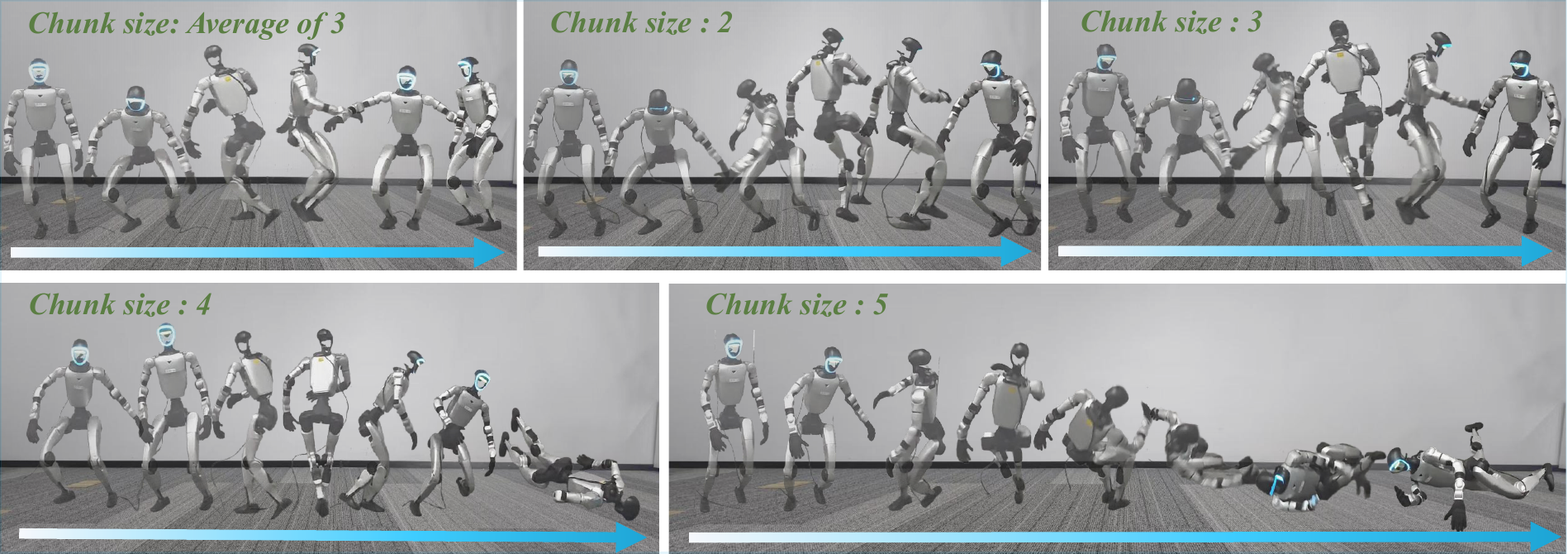}
\caption{Real-world performance of generating different action chunk sizes.}
\label{acs}
\end{figure*}

\paragraph{Different Action Chunk Sizes}
We present real-world visualization results for different action chunk sizes, using the identical experimental configuration to that reported for the quantitative results in the main text: during training, a random value of $k$ is sampled from the range [1, 3] to mask $\{o_{t-k+1}, ... t_t\}$. For inference, to generate an action chunk with a size of 3, we simply replace the current observation with the learned mask token [\text{M}], concatenate two [\text{M}] tokens to its rear, and extract the last three actions output by the model for continuous step execution. As illustrated in Figure \ref{acs}, the real-robot performance remains highly stable when the output action chunk size $k \leq 3$ during inference, while the robot experiences falls when $k > 3$. Videos of real-robot results with different action chunk sizes are provided in the supplementary material ($k \in [1,3]$; the robot fails to perform stably when the chunk size is 4 or 5).

\begin{figure}[!htbp]
\centering
  \includegraphics[width=1.02\textwidth]{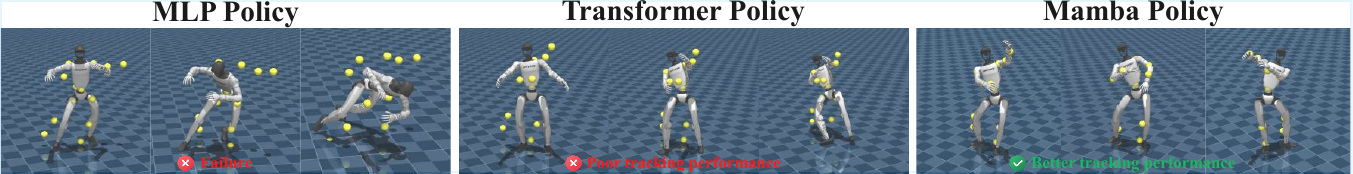}
\vspace{-2mm}
\caption{Qualitative results in the MuJoCo.}
\label{sim_app}
\vspace{-4mm}
\end{figure}

\paragraph{Simulation}
To validate the superiority of the Mamba policy, we visualize two simulation cases. As presented in the upper segment of Figure~\ref{sim_app}, the MLP policy demonstrates subpar tracking performance and the falling situation. Transformer policy also presents a poorer tracking performance. By contrast, our mamba policy attains enhanced tracking outcomes through its improved robustness and distribution modeling capacity.

\vspace{-2mm}

\begin{figure}[!htbp]
\centering
  \includegraphics[width=0.98\textwidth]{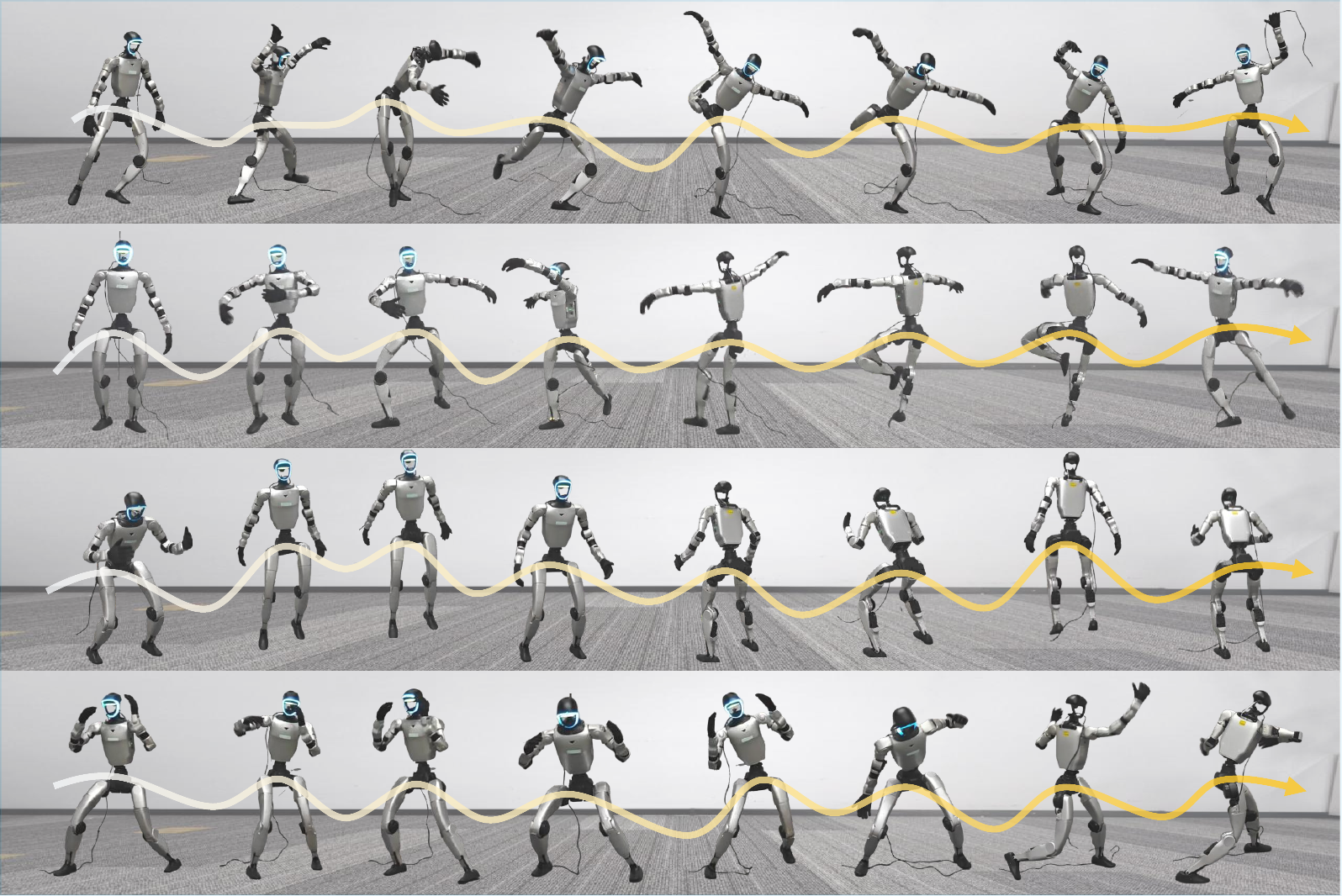}
\vspace{-2mm}
\caption{Real-world deployment.}
\label{fig:real_1}
\vspace{-4mm}
\end{figure}

\paragraph{Real-World}
We present real-world deployment as shown in Figures~\ref{fig:real_1} and~\ref{fig:real_2}. A supplementary video showcasing real-robot deployments is provided in the supplementary material.

\vspace{-2mm}

\vspace{-2mm}
\begin{figure}[!htbp]
\centering
  \includegraphics[width=0.98\textwidth]{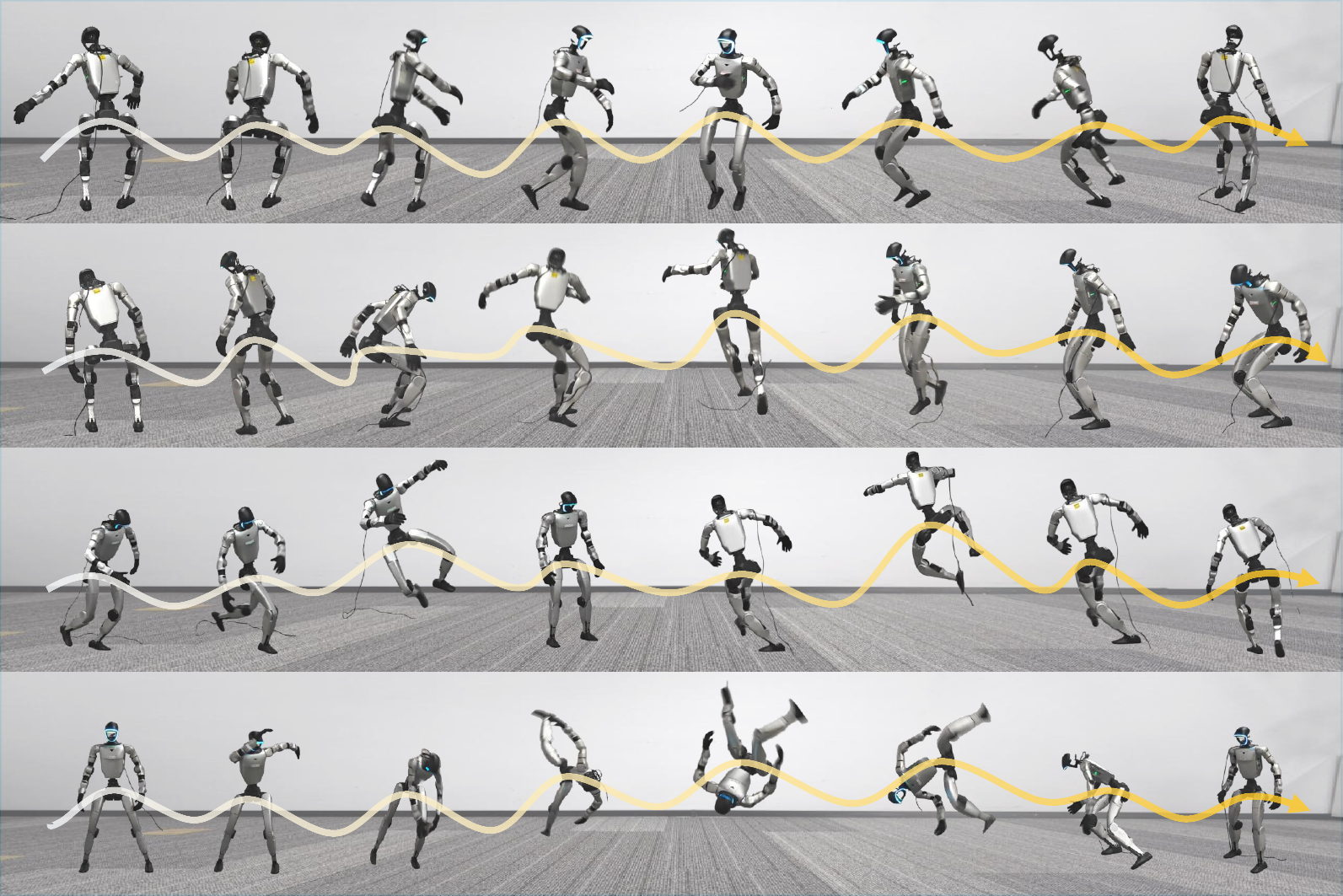}
\vspace{-2mm}
\caption{Real-world deployment.}
\label{fig:real_2}
\vspace{-4mm}
\end{figure}

\end{document}